\documentclass{article}

\usepackage[preprint]{neurips_2024}
\usepackage{amsmath,amsfonts,bm}
\usepackage{xspace,mathtools}

\usepackage{amsmath,amsfonts,bm}

\def\eqref#1{equation~\ref{#1}}

\def\1{\bm{1}}

\def\vzero{{\bm{0}}}

\def\ve{{\bm{e}}}

\def\vm{{\bm{m}}}
\def\vn{{\bm{n}}}

\def\vv{{\bm{v}}}

\def\vx{{\bm{x}}}
\def\vy{{\bm{y}}}
\def\vz{{\bm{z}}}

\def\mR{{\bm{R}}}

\def\mW{{\bm{W}}}

\DeclareMathAlphabet{\mathsfit}{\encodingdefault}{\sfdefault}{m}{sl}
\SetMathAlphabet{\mathsfit}{bold}{\encodingdefault}{\sfdefault}{bx}{n}

\def\gA{{\mathcal{A}}}

\def\gE{{\mathcal{E}}}

\def\gG{{\mathcal{G}}}

\def\gN{{\mathcal{N}}}

\def\gX{{\mathcal{X}}}

\def\gZ{{\mathcal{Z}}}

\newcommand{\R}{\mathbb{R}}

\DeclareMathOperator*{\argmax}{arg\,max}
\DeclareMathOperator*{\argmin}{arg\,min}

\DeclareMathOperator{\sign}{sign}

\newcommand{\airfrans}{AirfRANS\xspace}
\newcommand{\mlcfdLong}{NeurIPS 2024 ML4CFD Competition\xspace}

\newcommand{\score}{\textsc{Global Score}\xspace}
\newcommand{\physcore}{\textsc{Physics Score}\xspace}
\newcommand{\oodscore}{\textsc{OOD Score}\xspace}
\newcommand{\mlscore}{\textsc{ML Score}\xspace}
\newcommand{\propmp}{\textsc{Surf2Vol Message Passing}\xspace}
\newcommand{\sv}{\textsc{Surf2Vol}\xspace}

\newcommand{\mesh}{\gX}
\newcommand{\surfmesh}{\gX_{\operatorname{surf}}}
\newcommand{\freemesh}{\gX_{\operatorname{fs}}}
\newcommand{\foilmesh}{\gX_{\operatorname{af}}}
\newcommand{\downmesh}{\gX_{\operatorname{ds}}}
\newcommand{\trailmesh}{\gX_{\operatorname{trail}}}
\newcommand{\trailcoord}[1]{{#1}_{\operatorname{trail}}}
\newcommand{\trailx}[1]{{#1}_\tau}
\newcommand{\leadcoord}[1]{{#1}_{\operatorname{lead}}}
\newcommand{\drag}{C_D}
\newcommand{\lift}{C_L}
\newcommand{\inletvel}{\vv_{\infty}}

\newcommand{\cat}[1]{\left[#1\right]}
\newcommand{\norm}[1]{\lVert#1\rVert}
\newcommand{\velx}{\bar{u}_x}
\newcommand{\vely}{\bar{u}_y}
\newcommand{\pres}{\bar{p}}
\newcommand{\logpres}{\bar{q}}
\newcommand{\visc}{\nu_t}
\newcommand{\mlp}[1]{\operatorname{MLP}(#1)}
\newcommand{\gnn}[1]{\operatorname{GNN}(#1)}
\newcommand{\neighb}[1]{\gN\left(#1\right)}
\newcommand{\svneighb}[1]{\gN_{\operatorname{s2v}}(#1)}

\newcommand{\compo}{x}
\newcommand{\compl}{y}
\newcommand{\vect}[2]{\begin{bmatrix} {#1} & {#2} \end{bmatrix}^\top}
\newcommand{\ang}[1]{\theta(#1)}
\newcommand{\angf}[1]{\theta^4({#1})}
\newcommand{\rtn}[1]{\mR_{{#1}^\circ}}
\newcommand{\inrtn}{\mR_{\vv}}
\DeclareMathOperator*{\atan}{atan2}
\DeclareMathOperator*{\pefun}{PE}
\newcommand{\pe}[1]{\pefun({#1})}
\newcommand{\nbasis}{n_{\operatorname{basis}}}
\newcommand{\yl}[2]{Y^0_{#1}(#2)}
\newcommand{\oddyl}[2]{\bar{Y}^0_{#1}(#2)}
\DeclareMathOperator*{\sphfun}{SpH}
\newcommand{\sph}[1]{\sphfun(#1)}

\newcommand{\nodefeature}[2]{\vz_{#1}(#2)}
\newcommand{\nodeemb}[1]{\nodefeature{}{#1}}
\newcommand{\surfnodeemb}[1]{\nodefeature{\operatorname{surf}}{#1}}
\newcommand{\surfnodefun}{\vz_{\operatorname{surf}}}
\newcommand{\basenodefeatures}[1]{\nodefeature{\operatorname{base}}{#1}}
\newcommand{\innodefeatures}[1]{\nodefeature{\operatorname{in}}{#1}}
\newcommand{\trailnodefeatures}[1]{\nodefeature{\operatorname{trail}}{#1}}

\newcommand{\angnodefeatures}[1]{\nodefeature{\operatorname{polar}}{#1}}
\newcommand{\sinenodefeatures}[1]{\nodefeature{\operatorname{sine}}{#1}}
\newcommand{\sphnodefeatures}[1]{\nodefeature{\operatorname{sph}}{#1}}
\newcommand{\canonodefeatures}[1]{\nodefeature{\operatorname{inlet}}{#1}}

\newcommand{\edgefeature}[3]{\ve_{#1}(#2,#3)}

\newcommand{\svedgeemb}[2]{\edgefeature{\operatorname{s2v}}{#1}{#2}}
\newcommand{\baseedgefeatures}[2]{\edgefeature{\operatorname{base}}{#1}{#2}}
\newcommand{\inedgefeatures}[2]{\edgefeature{\operatorname{in}}{#1}{#2}}
\newcommand{\angedgefeatures}[2]{\edgefeature{\operatorname{polar}}{#1}{#2}}
\newcommand{\sineedgefeatures}[2]{\edgefeature{\operatorname{sine}}{#1}{#2}}
\newcommand{\sphedgefeatures}[2]{\edgefeature{\operatorname{sph}}{#1}{#2}}
\newcommand{\canonedgefeatures}[2]{\edgefeature{\operatorname{inlet}}{#1}{#2}}

\newcommand{\svmodel}{\sv}
\newcommand{\svgnnmodel}{\textsc{Surf2Vol+GNN}\xspace}
\newcommand{\trailmodel}{\textsc{Trail}\xspace}
\newcommand{\polarmodel}{\textsc{Polar}\xspace}
\newcommand{\sinemodel}{\textsc{Sine}\xspace}
\newcommand{\sphmodel}{\textsc{SpH}\xspace}
\newcommand{\canonmodel}{\textsc{Inlet}\xspace}
\newcommand{\geompnnmodel}{\textsc{GeoMPNN}\xspace}

\usepackage[utf8]{inputenc} 
\usepackage[T1]{fontenc}    
\usepackage{hyperref}       
\usepackage{url}            
\usepackage{booktabs}       
\usepackage{amsfonts}       
\usepackage{nicefrac}       
\usepackage{microtype}      
\usepackage{xcolor}         
\usepackage{subcaption}

\title{A Geometry-Aware Message Passing Neural Network for Modeling Aerodynamics over Airfoils}

\author{%
  Jacob Helwig \quad Xuan Zhang \quad Haiyang Yu \quad Shuiwang Ji \\
  Department of Computer Science \& Engineering 
  \\ Texas A\&M University \\
  \texttt{\{jacob.a.helwig,xuan.zhang,haiyang,sji\}@tamu.edu} 
}

\usepackage[textsize=tiny, disable]{todonotes}
\usepackage{regexpatch}
\makeatletter
\presetkeys{todonotes}{inline,caption={}}{}
\xpatchcmd{\@todo}{\setkeys{todonotes}{#1}}{\setkeys{todonotes}{inline,#1}}{}{}
\makeatother
\usepackage[capitalise,noabbrev,nameinlink]{cleveref}
\NewDocumentCommand{\shuiwang}
{ mO{} }{\textcolor{blue}{\textsuperscript{\textit{Shuiwang Ji}}\textsf{\textbf{\small[#1]}}}}

\begin{document}

\maketitle

\begin{abstract}
    Computational modeling of aerodynamics is a key problem in aerospace engineering, often involving flows interacting with solid objects such as airfoils.
    Deep surrogate models have emerged as purely data-driven approaches that learn direct mappings from simulation conditions to solutions based on either simulation or experimental data.
    Here, we consider modeling of incompressible flows over solid objects, wherein geometric structures are a key factor in determining aerodynamics. 
    To effectively incorporate geometries, we propose a 
    message passing scheme that efficiently and expressively integrates the airfoil shape with the mesh representation.    
    Under this framework, we first obtain a representation of the geometry in the form of a latent graph on the airfoil surface. We subsequently propagate this representation to all collocation points through message passing on a directed, bipartite graph. We demonstrate that this framework supports efficient training by downsampling the solution mesh while avoiding distribution shifts at test time when evaluated on the full mesh. 
    To enable our model to be able to distinguish between distinct spatial regimes of dynamics relative to the airfoil, we represent mesh points in both a leading edge and trailing edge coordinate system. We further enhance the expressiveness of our coordinate system representations by embedding our hybrid Polar-Cartesian coordinates using sinusoidal and spherical harmonics bases. We additionally find that a change of basis to canonicalize input representations with respect to inlet velocity substantially improves generalization.
    Altogether, these design choices lead to a purely data-driven machine learning framework known as \textsc{GeoMPNN}, which won the Best Student Submission award at the \mlcfdLong, placing fourth overall. Our code is publicly available as part of the AIRS library (\url{https://github.com/divelab/AIRS}).
\end{abstract}

\section{Introduction}


The numerical solution of PDEs has many applications across a wide variety of disciplines. Due to the high cost of numerical simulation, machine learned surrogate models of dynamics have recently gained traction~\citep{lu2021learning, li2021fourier, tran2023factorized, helwig2023group, zhang2024sinenet}. Inverse design, wherein a physical system is optimized according to a design objective, has great potential for acceleration by neural surrogates due to the requirement of a many solver calls to evaluate candidate designs~\citep{zhang2023artificial}. A prominent example is airfoil design optimization, where each proposed airfoil geometry is evaluated under a variety of operating conditions according to metrics such as drag and lift~\citep{bonnet2022airfrans, yagoubi2024neurips}. In this work, we aim to accelerate airfoil design by developing neural surrogates trained to generalize over flow conditions and airfoil geometries. Due to the high influence of the airfoil shape on the resulting dynamics, it is not only key to learn an expressive representation of the airfoil shape, but to efficiently incorporate this representation into the predicted fields. Furthermore, to handle a large number of simulation mesh points, the design of architectures that are amenable to efficient training strategies is vital.

To tackle these challenges, we propose a geometry-aware message passing neural network that first extracts a geometric representation of the airfoil shape, and then predicts the surrounding field conditioned on this representation. For efficient training, the architecture is specifically designed to handle subsampled meshes during training without degradation in the solution accuracy on the full mesh at test time. We additionally propose a variety of physically-motivated enhancements to our coordinate system representation, including multiple coordinate systems centered at various key points of the airfoil, the use of both polar and Cartesian systems, and embeddings in sinusoidal and spherical harmonics bases. We furthermore find that a change of basis to canonicalize input representations with respect to inlet velocity substantially improves generalization. We validate our resulting framework, known as \textsc{GeoMPNN}, on the AirfRANS~\citep{bonnet2022airfrans} dataset in the \mlcfdLong~\citep{yagoubi2024neurips}. Our designs are proven to be effective through extensive experiments.

\section{Airfoil Design with the Reynolds-Averaged Navier Stokes Equations}


The \airfrans dataset \citep{bonnet2022airfrans} models the two-dimensional incompressible
steady-state Reynolds-Averaged Navier–Stokes (RANS) equations over airfoils of various shapes and under varying conditions with the purpose of accelerating airfoil design optimization. Given the geometry and dynamics parameters for a particular instantiation of the system, the task is to predict various physical quantities on all mesh points, enabling computation of forces acting on the airfoil. These forces are described by the drag $\drag$ and lift $\lift$ coefficients, and are important in determining the effectiveness of a particular airfoil design. 

In the \mlcfdLong~\citep{yagoubi2024neurips}, 103 solutions are provided as training examples, along with an in-distribution test set of 200 solutions and an out-of-distribution (OOD) test set of 496 solutions. In~\cref{sec:inputs}, we describe the inputs provided for each example, and in~\cref{sec:targets}, we discuss the fields to be predicted, as well as the means by which model performance is evaluated.   

\subsection{System Descriptors}\label{sec:inputs}

For each set of simulation conditions, the mesh coordinates are provided as Cartesian coordinates $\mesh\subset\R^2$ centered at the leading edge of the airfoil\footnote{In the original version of the \airfrans dataset, the mesh coordinate system places the origin approximately at the leading edge of the airfoil, which we define as the left-most point in $\surfmesh$. We re-orient the coordinate system to place the origin exactly at the leading edge.}, with the airfoil geometry described by the points on the surface of the airfoil $\surfmesh\subset\mesh$. Throughout the following sections, we refer to $\surfmesh$ as the \emph{surface mesh}, and $\mesh$ as either the \emph{mesh} or the \emph{volume mesh}. For $\vx\in\surfmesh$, the function $\vn:\mesh\to\R^2$ returns the surface normal for each point on the airfoil surface, and returns the null vector $\vn(\vx')=\vzero$ for $\vx'\in\mesh\setminus\surfmesh$ not on the surface of the airfoil. In addition to the airfoil geometry, simulation conditions are defined by the inlet velocity vector $\inletvel\in\R^2$. The distance function $d:\mesh\to\R$ maps $\vx$ to its orthogonal distance to the airfoil surface $\surfmesh$. We augment the base features to include distance to the leading edge of the airfoil $\norm{\vx}$. The overall base input features for each mesh point are given by
\begin{equation}\label{eq:basefeats}
    \basenodefeatures\vx\coloneqq\cat{\vx,\vn(\vx),\inletvel, d(\vx),\norm{\vx}}\in\R^8.
\end{equation}

\subsection{Modeling Task and Evaluation}\label{sec:targets}
\todo{Add visualization of fields}

Given inputs $\basenodefeatures \vx$ for the mesh $\mesh$, solving the steady-state RANS equations entails predicting the steady-state fluid velocity in the $x$ direction $\velx$ and $y$ direction $\vely$, the steady-state reduced pressure $\pres$, and the steady-state turbulent viscosity $\visc$\citep{bonnet2022airfrans}. The value of these functions must be predicted for each $\vx\in\mesh$.
Given these quantities, the lift $\lift$ and drag $\drag$ coefficients can be computed. We report errors for normalized fields, where the normalization is such that the field has mean 0 and standard deviation 1 across the training set.

The \mlcfdLong follows the LIPS framework~\citep{leyli2022lips} to evaluate model performance. Models are assessed along three dimensions, resulting in an overall \score out of 100. On the in-distribution test set, an in-distribution ML-related performance score (\mlscore) is determined in terms of the degree of error across all 200 examples on these four variables, as well as pressure $\pres$ on the surface $\surfmesh$, and computation time. To assign an in-distribution physics compliance score (\physcore), $\lift$ and $\drag$ are computed using the predictions and compared to the ground truth. An out-of-distribution generalization score (\oodscore) is assigned by computing both the ML-related performance score and the physics compliance score on the 496 out-of-distribution examples. The \score is calculated by taking a weighted average across \mlscore, \physcore, and \oodscore. 

\section{Learned Simulation Pipeline}

In~\cref{sec:trainProc}, we describe training procedures, while in~\cref{sec:baselines}, we introduce our baseline models.

\subsection{Training Procedures}\label{sec:trainProc}

\begin{figure}
    \centering
    \includegraphics[width=\linewidth]{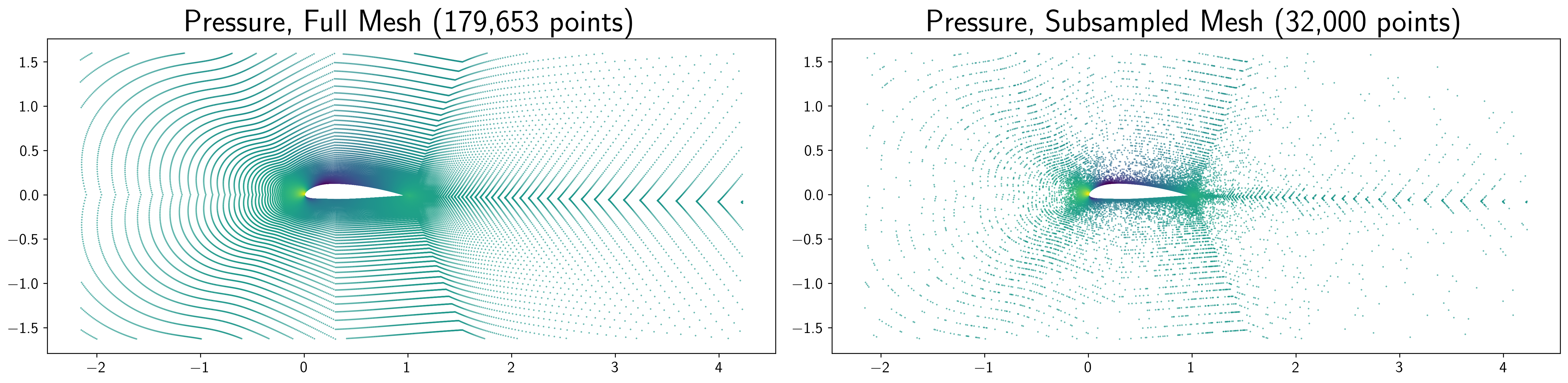}
    \caption{Full mesh (\textit{left}) and subsampled mesh (\textit{right}). To reduce training costs, we randomly sample 32K mesh points for each training example at each epoch. Evaluation is conducted at the full resolution. }
    \label{fig:subsamp}
\end{figure}

All models are implemented in PyTorch~\citep{paszke2019pytorch} and PyTorch Geometric~\citep{Fey/Lenssen/2019} and are optimized using the Adam optimizer~\citep{adamKingmaB14} using the 1cycle learning rate scheduler~\citep{smith2019super} with a maximum learning rate of $1\times10^{-3}$ for 600 epochs with a batch size of 1. As the number of mesh points $\lvert\mesh\rvert$ is greater than 179K on average, it is inefficient to supervise the solution on the entirety of the mesh points on each training update~\citep{bonnet2022airfrans}. Instead, we randomly sample 32K points from each mesh every epoch and only supervise the solution on these points, as shown in~\cref{fig:subsamp}. During inference, predictions are made on the full mesh. For all experiments, we train 1 model per field $\velx,\vely,\pres$, and $\visc$, resulting in 4 total models per experiment. We repeat each experiment 8 times.

\subsection{Baseline Methods}\label{sec:baselines}
As baselines, we train an MLP and GNN, both per field. The MLP is applied point-wise over the mesh as
\begin{equation*}
    \left\{\mlp{\innodefeatures{\vx}}\right\}_{\vx\in\mesh},   
\end{equation*}
where we set $\innodefeatures\vx=\basenodefeatures\vx$. 
We denote the input graph for the GNN by $\gG(\gZ(\mesh), \gA(\mesh), \gE(\mesh))$, where 
\begin{equation}\label{eq:gnnnodefeats}
    \gZ(\mesh)\coloneqq\{\innodefeatures\vx\}_{\vx\in\mesh}    
\end{equation}
are the input node features, $\gA(\mesh)$ is the edge set, and $\gE(\mesh)$ are the input edge features. The edge set $\gA(\mesh)$ is constructed via a radius graph with radius $r=0.05$ such that the neighborhood of $\vx\in\mesh$ is given by 
\begin{equation}\label{eq:radGr}
    \neighb\vx\coloneqq\{\vy:\vy\in\mesh,\norm{\vx-\vy}\leq r\}.
\end{equation}
\todo{Add sweep of number of neighbors}
If the number of neighbors $\lvert\neighb\vx\rvert$ exceeds a threshold $M$, $M$ neighbors are sampled randomly from $\gN(\vx)$. Here, we set $M=4$. The base edge features for $\vx\in\mesh$ and $\vy\in\neighb\vx$ are given by
\begin{equation*} 
    \baseedgefeatures\vy\vx\coloneqq \cat{\vy-\vx,\norm{\vy-\vx}}\in\R^3.
\end{equation*}
Again with $\inedgefeatures\vx\vy=\baseedgefeatures\vx\vy$, the input edge features are then given by
\begin{equation}\label{eq:baseEdgeFeatMatrix}
    \gE(\mesh)\coloneqq\{\inedgefeatures\vx\vy:\vx\in\mesh, \vy\in\neighb\vx\}.
\end{equation}

\section{Methods and Results}

\begin{figure}
    \centering
    \includegraphics[width=\linewidth]{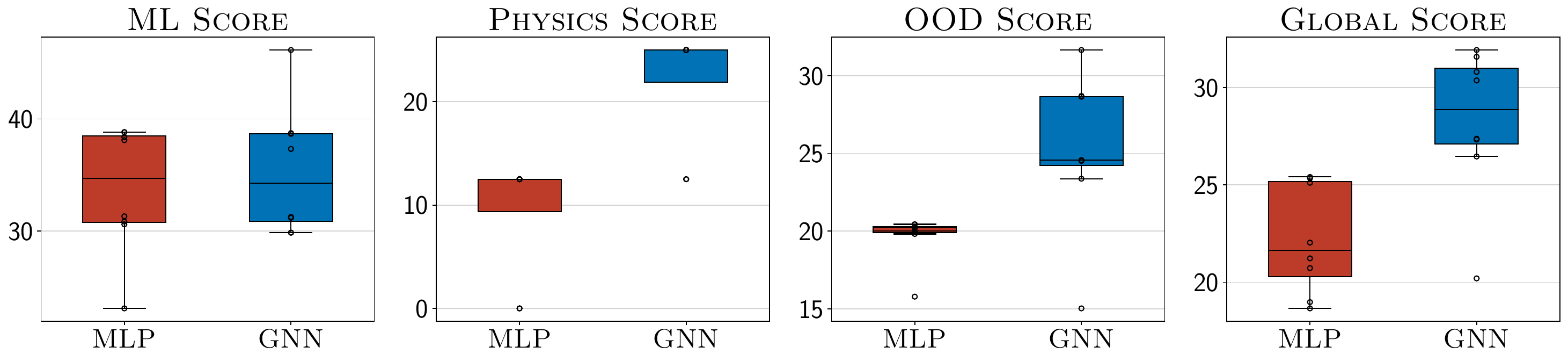}
    \caption{Scores for the MLP and GNN baselines across 8 runs.}
    \label{fig:gnn_mlp_scores}
\end{figure} 

The top performing method from the previous iteration of the competition achieved a \score of $47.04$\footnote{\cite{yagoubi2024neurips} report the score of \cite{casenave2024mmgp} as $81.29$, however, under the updated scoring for the current iteration of the competition, their score is $47.04$.} using Gaussian processes~\citep{casenave2024mmgp}. As can be seen from~\cref{fig:gnn_mlp_scores}, both the MLP and GNN underperform~\cite{casenave2024mmgp}. We further analyze the performance of baselines in~\cref{sec:baselineperf} and use these insights to devise an approach for more effectively encoding the geometry of the airfoil in~\cref{sec:stov}. We explore augmenting the coordinate system of the mesh in~\cref{sec:coords} and enhance the expressiveness of this representation with coordinate system embeddings in~\cref{sec:coordEmb}. To canonicalize input representations with respect to the direction of the inlet velocity, we introduce a change of basis in~\cref{sec:canon}. To handle extreme values in the pressure field, we apply a log transform to targets as described in~\cref{sec:logPres}.


\subsection{Baseline Analysis}\label{sec:baselineperf}
\todo{Add ablation on number of neighbors for GNN}
We next analyze the performance, weaknesses, and strengths of the baseline methods to motivate our proposed \propmp. 

\subsubsection{Airfoil Geometry Encoding} 

As the MLP is applied point-wise and does not model any spatial interactions, it does not effectively encode the shape of the airfoil, despite the importance of shape in determining the resulting solution fields. As can be seen in~\cref{fig:gnn_mlp_scores}, the GNN improves both the \physcore and \oodscore relative the MLP, which could be due to the ability of message passing to encode the shape of the airfoil via message passing between the surface mesh points $\surfmesh$. However, the GNN does not substantially improve the \mlscore relative to the MLP, for which we offer two potential explanations. 

First, there does not exist a dedicated mechanism for encoding the geometry, as parameters in message passing layers are simultaneously used to update representations for mesh points not on the surface, that is, $\vx\in\mesh\setminus\surfmesh$. This sharing likely limits the expressiveness of the airfoil representation. Second, due to the locality of message passing~\citep{gilmer2017neural,battaglia2018relational}, any representation of the airfoil shape which the model manages to learn cannot be propagated beyond the few-hop neighbors of the surface mesh points. Although the solution values on mesh points outside of this radius are affected by the airfoil geometry, the ability of the GNN to propagate this information across the mesh is limited, and therefore, these relationships are not effectively modeled. 

\subsubsection{Increased Evaluation Resolution}\label{sec:incRes}

\begin{figure}
    \centering
    \includegraphics[width=\linewidth]{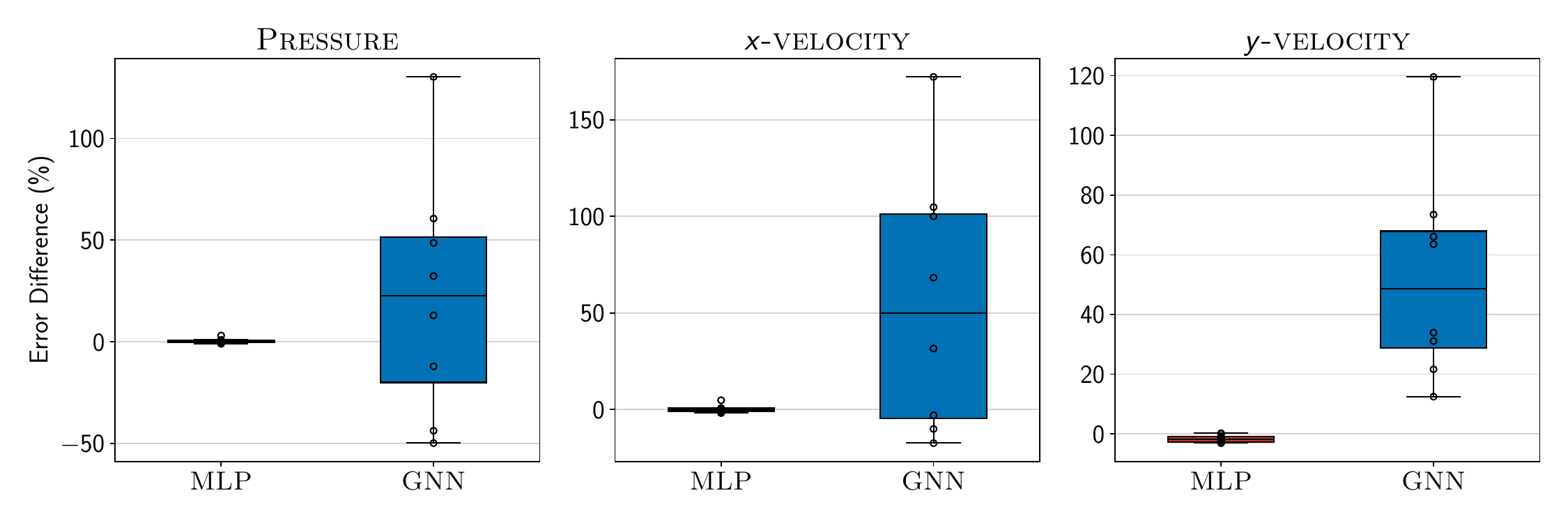}
    \caption{Difference in errors between subsampled and full resolution for MLP and GNN. We analyze the effect of increasing evaluation resolution by assessing the relative difference in the test error on the full resolution compared to the error on the subsampled resolution, defined in~\cref{eq:reldiff}. As can be seen, increasing resolution substantially increases the error on each field for the GNN. Note that error on the turbulent viscosity field is omitted due to overfitting for both the MLP and GNN.}
    \label{fig:gnnerrdiff}
\end{figure}

As discussed in~\cref{sec:trainProc}, due to the large number of mesh points, the mesh is downsampled during training, but remains at the original resolution during evaluation. To assess the effect of increasing the resolution, we compare the error on each field at the full resolution $\varepsilon_{\operatorname{full}}$ and at the randomly subsampled resolution $\varepsilon_{32\text{K}}$. In~\cref{fig:gnnerrdiff}, we compute a relative difference defined by
\begin{equation}\label{eq:reldiff}
    \frac{\varepsilon_{\operatorname{full}} - {\varepsilon_{32\text{K}}}}{\varepsilon_{32\text{K}}}.
\end{equation}
While the MLP is applied point-wise and therefore experiences no distribution shift, we can see a substantial degradation in the performance of the GNN on the subsampled mesh versus on the full-resolution mesh, possibly due to a shift in the neighborhood structure. \cite{bonnet2022airfrans} suggest repeatedly sampling the mesh and applying the GNN in a loop until all mesh points have been sampled during evaluation. However, our initial attempts showed that a high number of forward passes of the model were needed to cover all mesh points, and sampling without replacement between iterations led to an even more severe distribution shift in the neighborhood structure. This further motivates us to look beyond generic message passing algorithms for this problem.

\subsection{Geometry Encoding with \propmp}\label{sec:stov}

\begin{figure}
    \centering
    \includegraphics[width=\linewidth]{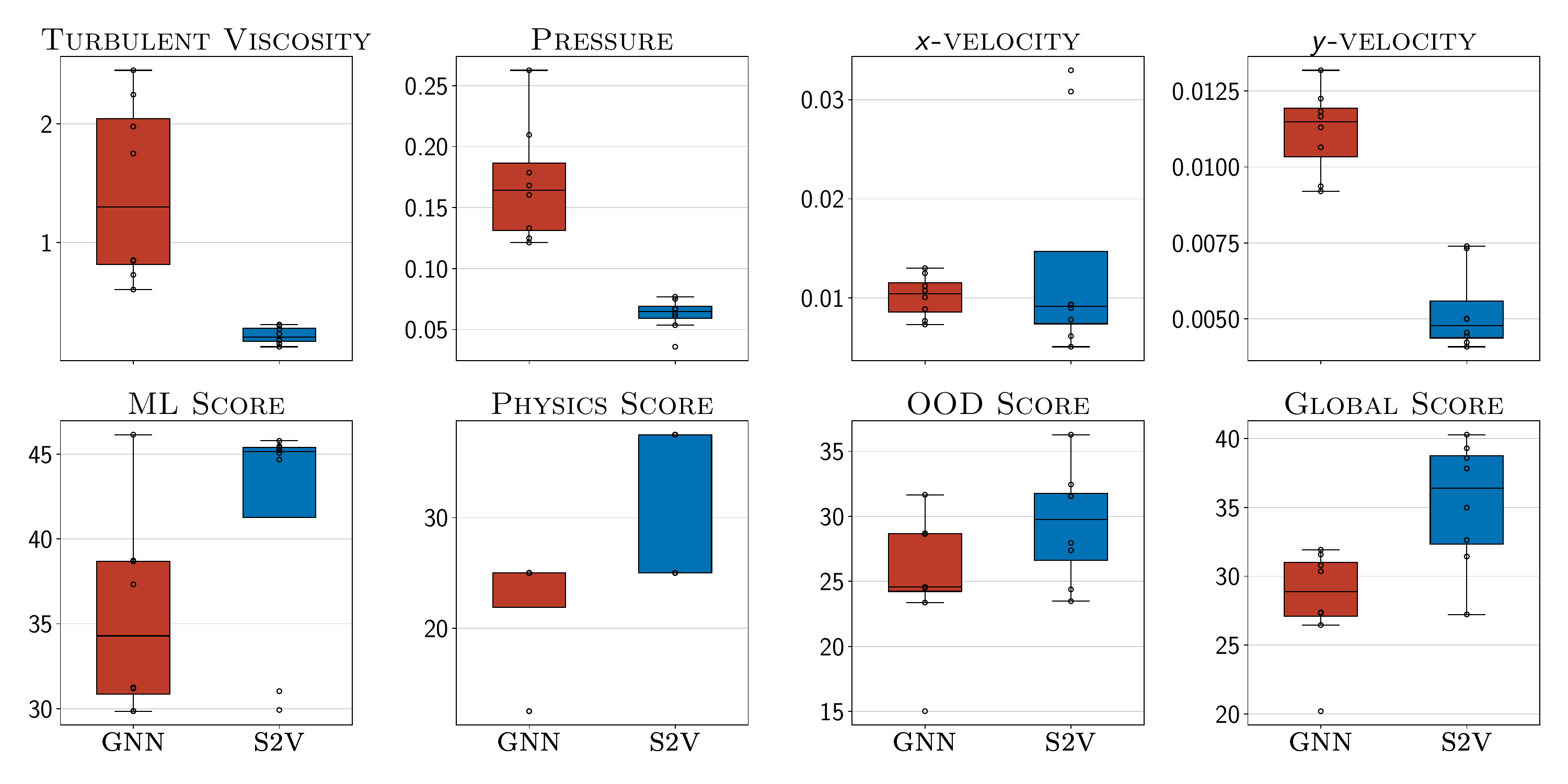}
    \caption{Comparison of the \svmodel model to the GNN. \textit{Top}: MSE errors on each field, where lower is better, and the targets are normalized as discussed in~\cref{sec:targets}. \textit{Bottom}: Scores in each category, where higher is better.}
    \label{fig:sv_v_gnn_res}
\end{figure}

Our analysis in~\cref{sec:baselineperf} suggests that to improve model performance, we need to retain the ability of the GNN to encode the airfoil geometry, and furthermore integrate dedicated mechanisms for representing this shape and subsequently distributing this representation across the mesh. Simultaneously, we would like to engrain the ability of the MLP to avoid error degradation when transferring from the subsampled training meshes to the full-resolution evaluation meshes to avoid test-time distribution shifts.

With these design goals in mind, we propose \propmp. Under this scheme, we obtain a latent graph representation of the airfoil geometry. We then associate every point $\vx\in\mesh$ with a neighborhood $\svneighb\vx\subset\surfmesh$ of the latent graph using directed \sv edges from $\vy\in\svneighb\vx$ to $\vx$, thereby forming a directed bi-partite graph. The global propagation of the latent airfoil representation is then facilitated by the \sv edges.

\todo{Check that number of neighbors is same}
The latent graph representation of the airfoil geometry is obtained via $L=4$ layers of learned message passing over the mesh points $\vy\in\surfmesh$ following a standard message passing scheme~\citep{gilmer2017neural,battaglia2018relational}. The input graph $\gG(\gZ(\surfmesh), \gA(\surfmesh), \gE(\surfmesh))$ is formed analogously to~\cref{eq:gnnnodefeats,eq:baseEdgeFeatMatrix,eq:radGr} using radius $r=0.05$ and threshold $M=8$, and the final output of this geometric encoding is a latent representation $\surfnodeemb\vy$ for all $\vy\in\surfmesh$ as
\begin{equation}\label{eq:surfmp}
    \surfnodefun\coloneq\gnn{\gZ(\surfmesh), \gA(\surfmesh), \gE(\surfmesh)}.
\end{equation}
To propagate this representation across the mesh, we first form the \sv neighborhood $\svneighb\vx$ for mesh point $\vx$ using $k$-nearest neighbors with $k=8$ such that
\begin{equation*}
    \svneighb\vx\coloneqq\{\vy:\vy\in\surfmesh,\norm{\vx-\vy}\leq r_k(\vx)\},    
\end{equation*}
where $r_k(\vx)$ is set such that $\lvert\svneighb\vx\rvert=k$. 
For $\vy\in\svneighb{\vx}$, node features and \sv edge features are embedded as
\begin{align*}
       \nodeemb\vx\coloneq\mlp{\innodefeatures\vx} 
       &&
       \svedgeemb\vy\vx\coloneq\mlp{\inedgefeatures{\vy}{\vx}}.
\end{align*}
In each message-passing layer, edge embeddings are first updated as 
\begin{align*}
    \svedgeemb{\vy}{\vx}\leftarrow\svedgeemb{\vy}{\vx} + \mlp{\cat{\surfnodeemb{\vy}, \nodeemb{\vx},\svedgeemb{\vy}{\vx}}},
\end{align*}
thereby integrating a representation of the airfoil geometry in the form of the surface node embeddings $\surfnodeemb{\vy}$ closest to $\vx$ into the edge embedding $\svedgeemb{\vy}{\vx}$. This geometric representation is then aggregated into a message $\vm(\vx)$ via mean aggregation as 
\begin{equation*}
    \vm(\vx)\coloneq \frac1k\sum_{\vy\in\svneighb{\vx}}\svedgeemb{\vy}{\vx}.
\end{equation*}
As $\vm(\vx)$ aggregates representations of the airfoil from $\vy\in\surfmesh$ closest to $\vx$, it can be regarded as an embedding of the region of the airfoil geometry closest to $\vx$. The local geometric embedding is then integrated into the node embedding of $\vx$ as 
\begin{equation*}
    \nodeemb\vx\leftarrow\nodeemb\vx+\mlp{\cat{\nodeemb\vx,\vm(\vx)}}.
\end{equation*}

Following the final message passing layer, we apply a decoder MLP to the final latent node representations $\nodeemb\vx$ to obtain the predicted field at mesh point $\vx$. We assess these predictions in~\cref{fig:sv_v_gnn_res}, where we can observe that the \svmodel model substantially improves all metrics compared to the GNN. 

\subsubsection{Increased Evaluation Resolution and Spatial Interactions}

\begin{figure}
    \centering
    \includegraphics[width=\linewidth]{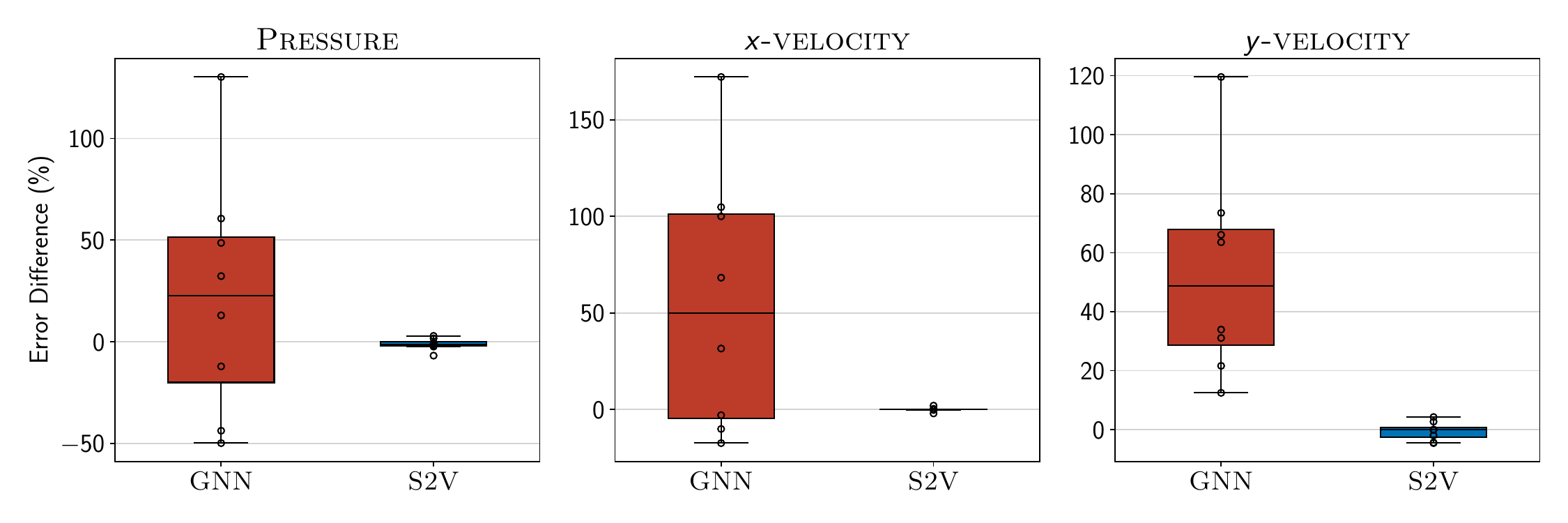}
    \caption{Difference in errors between subsampled and full resolution for \svmodel model and GNN. The \svmodel model error does not change with increased resolution.}
    \label{fig:errDiffSV}
\end{figure}
\begin{figure}
    \centering
    \includegraphics[width=\linewidth]{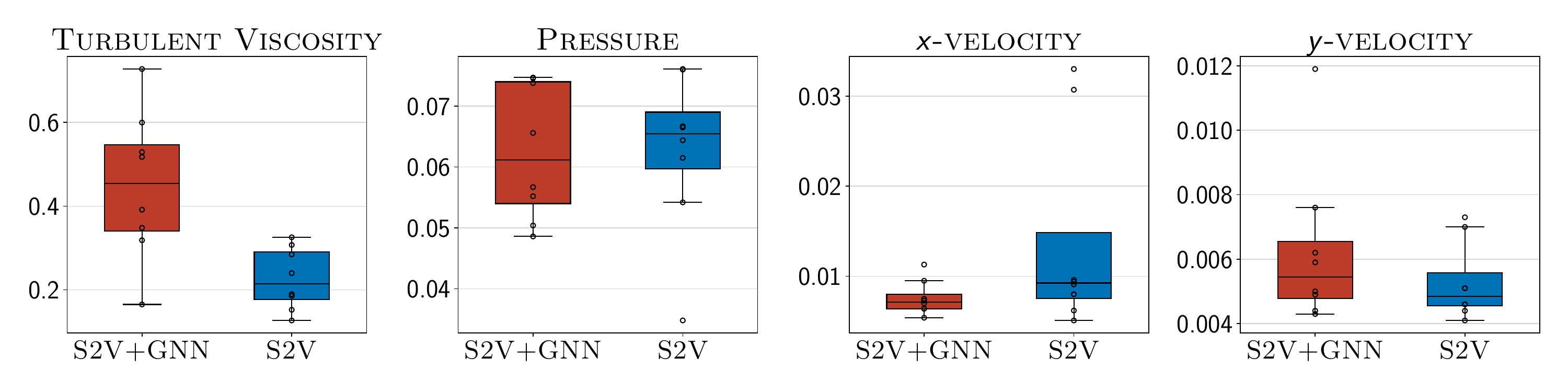}
    \caption{Results for the \svmodel model compared to the \svgnnmodel on the subsampled test set. Although the \svgnnmodel adds additional operations in the form of message passing between volume nodes following each \propmp layer, it does not offer a substantial improvement in performance on any of the fields.}
    \label{fig:svgnnres}
\end{figure}

As discussed in~\cref{sec:incRes}, the GNN demonstrates substantial degradation when the number of mesh points is increased. To avoid a distribution shift with the \svmodel model, during training, we do not subsample the nodes of the input surface graph  $\gG(\gZ(\surfmesh), \gA(\surfmesh), \gE(\surfmesh))$ for obtaining the geometric representation in~\cref{eq:surfmp}. As there are less than $1{,}011$ nodes on the airfoil surface on average, the cost of maintaining the full surface graph is minimal. While we still subsample the volume mesh $\mesh$, there are no interactions between pairs of volume nodes. This avoids any shifts in the neighborhood structure for the volume node $\vx$, as because we do not downsample the surface graph, the \sv neighborhood $\svneighb\vx$ is unchanged between training and evaluation. In~\cref{fig:errDiffSV}, we empirically verify that the error on the full and subsampled test set is roughly equal for the \svmodel model.

One of the main ways the \svmodel model avoids distribution shift with increased resolution is by only modeling the interaction between pairs of surface nodes and the directed interaction from surface nodes to volume nodes. However, this does not account for interactions between pairs of volume nodes. To assess the importance of volume node interactions, we add a volume message passing layer between each \propmp layer to form the \svgnnmodel model. To remove the effect of any potential distribution shift, we compare the \svgnnmodel model to the \svmodel model on the test set randomly subsampled to the same resolution as the training set. As can be seen from the results in~\cref{fig:svgnnres}, despite the additional operations, this modification does not offer a substantial benefit for any of the fields. This suggests that the most important interactions in this problem are surface-surface and surface-volume. 

To understand the limited benefit of volume-volume interactions, note that in the present problem, the target solution is steady-state. This is in contrast to time-evolving dynamical systems, where spatial processing mechanisms such as convolutions play a key role in modeling physical phenomena such as advection and diffusion~\citep{gupta2023towards,zhang2024sinenet}. Alternatively, steady-state problems featuring an input field  such as the coefficient function in Darcy Flow~\citep{li2021fourier} may similarly benefit from spatial processing mechanisms such as message passing~\citep{li2020neural} or attention~\citep{cao2021choose,li2023transformer,hao2023gnot}. However, as can be seen from~\cref{eq:basefeats}, the input fields featured in the present problem are primarily positional in nature, \textit{e.g.,} coordinates and distances, whereas the inlet velocity $\inletvel$ is a global feature shared by all mesh points. We therefore suggest that spatial processing mechanisms enabling communication between adjacent volume mesh points may be of limited benefit in such problems.



\todo{Add \sv fig}

\todo{Add \svmodel results (train + eval)}

\subsection{Augmented Coordinate System}\label{sec:coords}

In this section, we describe enhancements to our coordinate system representation to more faithfully represent relations between the airfoil geometry and mesh regions in which interactions with the airfoil give rise to distinct patterns of dynamics. In~\cref{sec:trailCoords}, we introduce a trailing edge coordinate system, while in~\cref{sec:polarCart}, we discuss the addition of angles to input features. 

\subsubsection{Trailing Edge Coordinate System}\label{sec:trailCoords}

\begin{figure}
    \centering
    \includegraphics[width=0.5\linewidth]{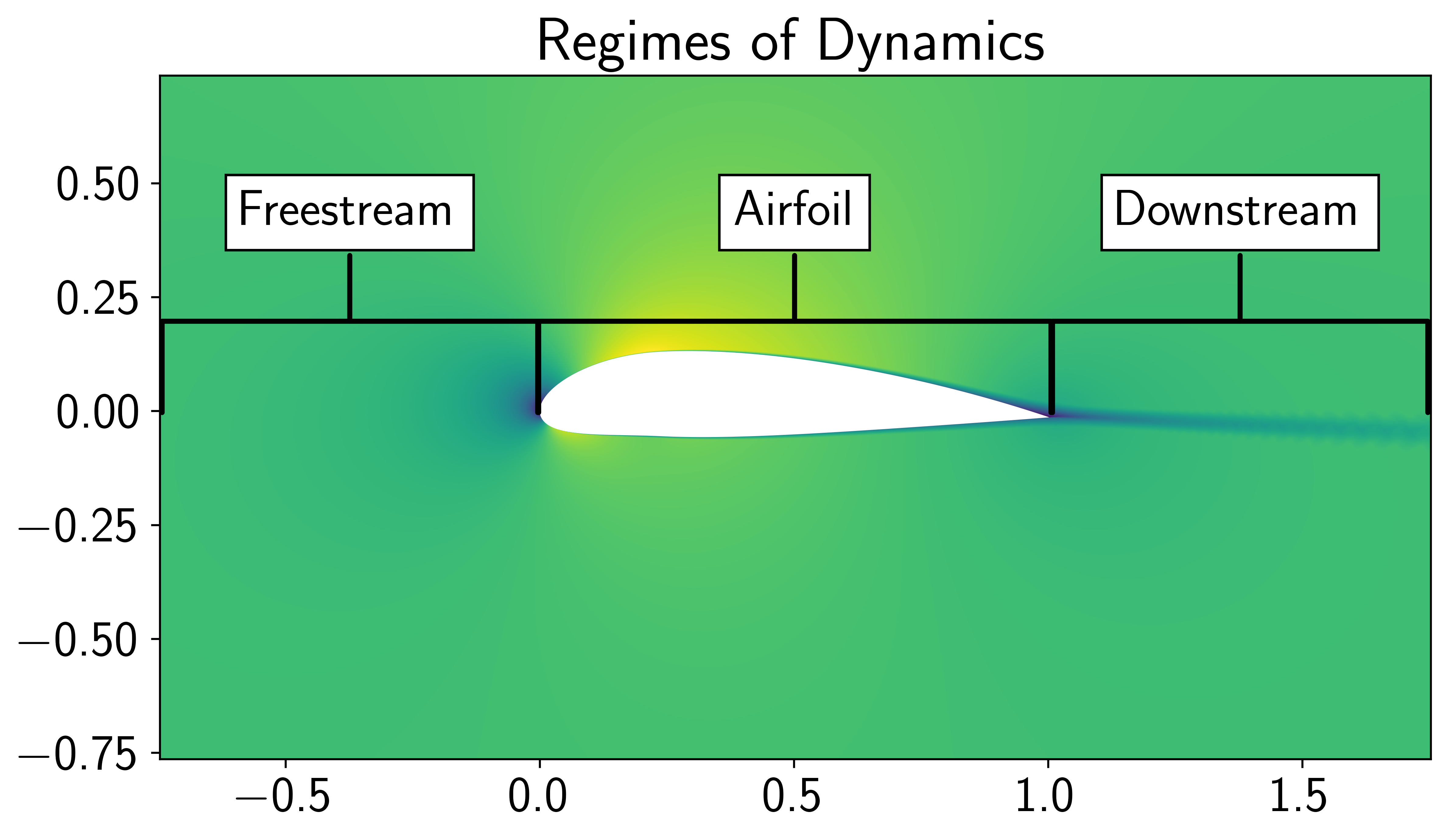}
    \caption{Spatial regimes of dynamics. Interactions between the airfoil and flow are unique in each of these three regions, creating distinct patterns of dynamics. Therefore, coordinate system representations should enable the model to distinguish between each of these regions. }
    \label{fig:dynRegimes}
\end{figure}

\begin{figure}
\centering
\begin{subfigure}{.5\textwidth}
  \centering
  \includegraphics[width=\linewidth]{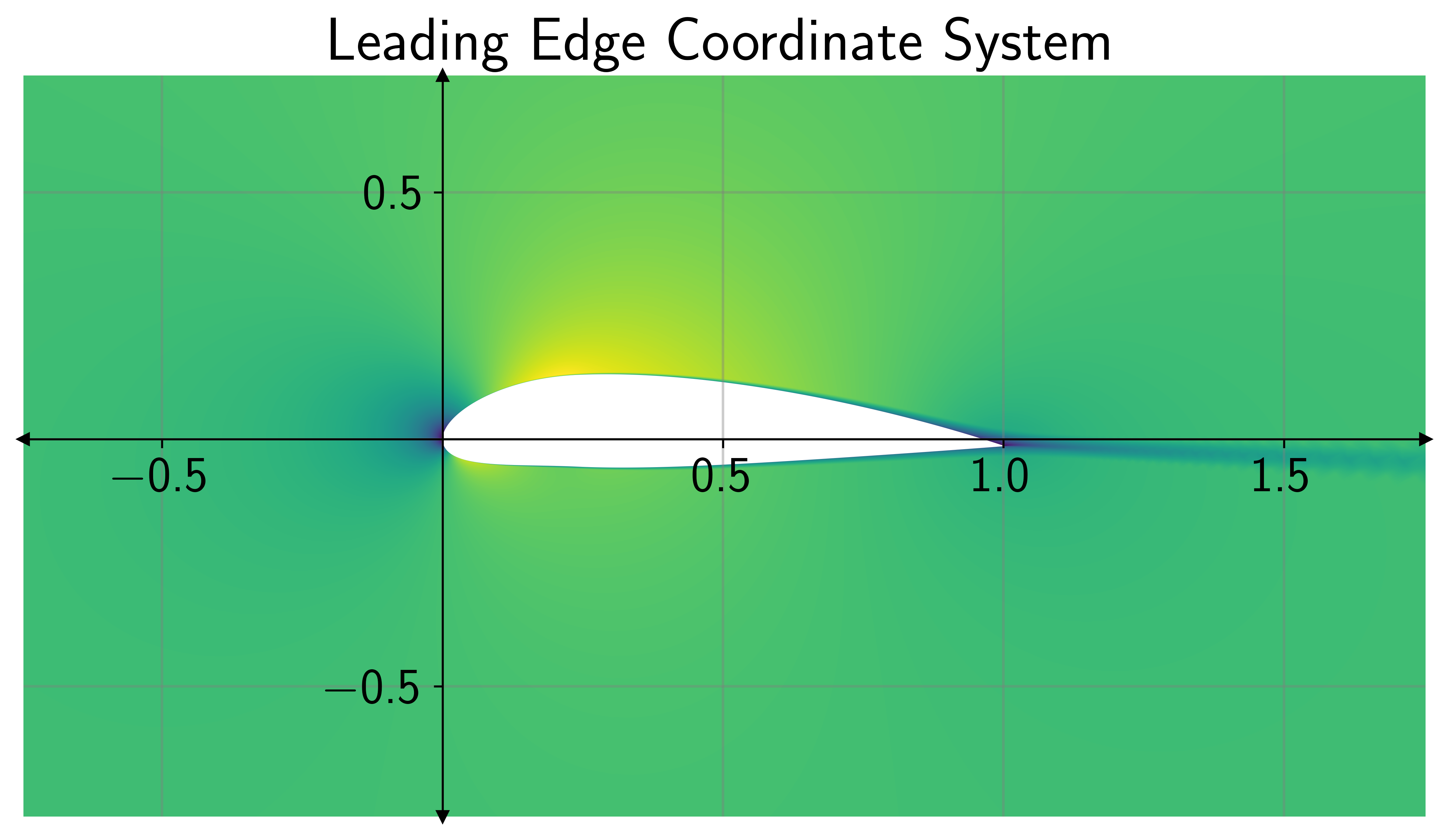}
  \caption{Leading edge coordinate system.}
  \label{fig:leadingEdge}
\end{subfigure}%
\begin{subfigure}{.5\textwidth}
  \centering
  \includegraphics[width=\linewidth]{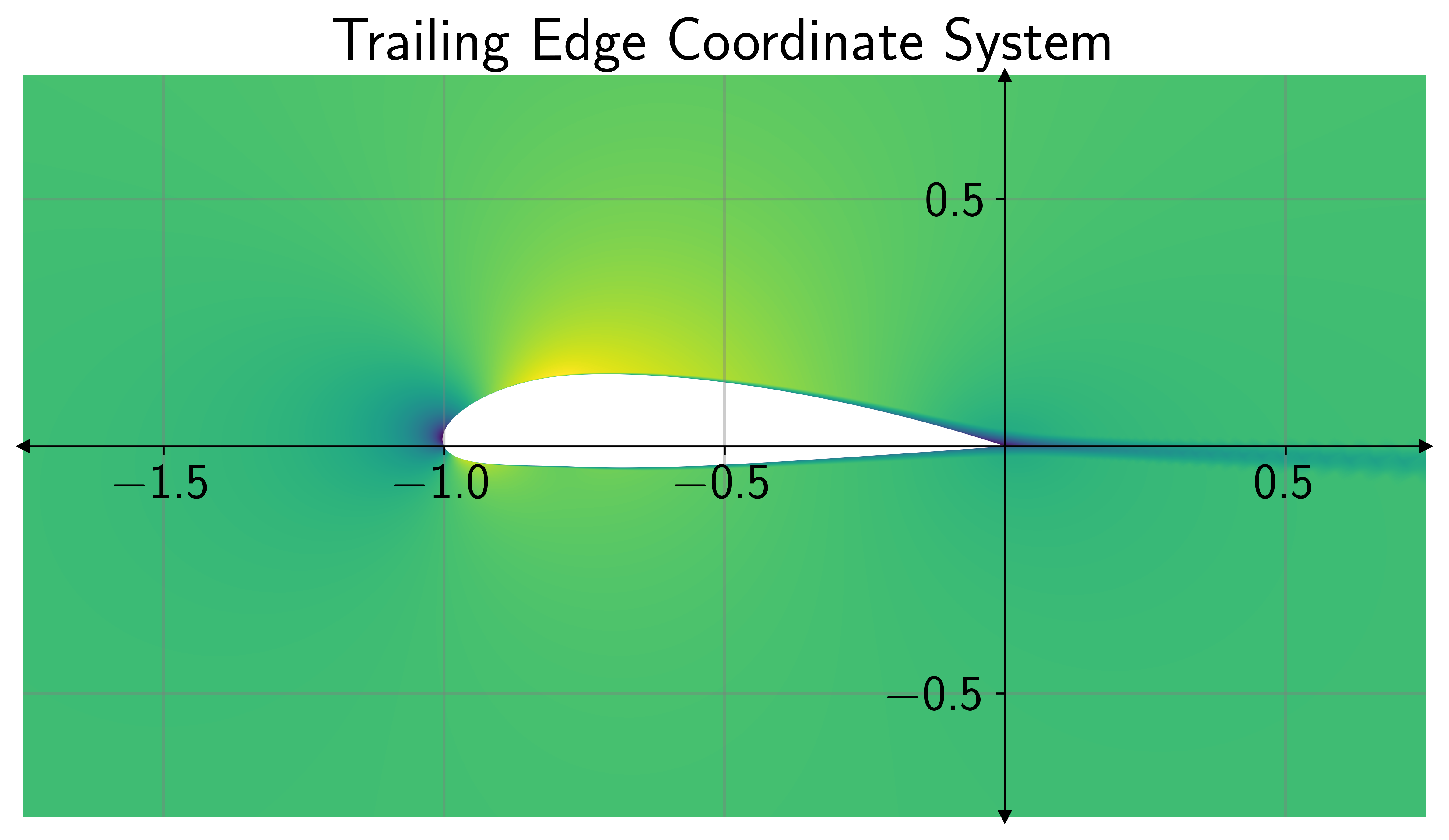}
  \caption{Trailing edge coordinate system.}
  \label{fig:trailingEdge}
\end{subfigure}
\caption{Leading and trailing edge coordinate systems. The leading and trailing edge coordinate systems enable the model to be able to distinguish between the three regimes of dynamics described in~\cref{fig:dynRegimes}.}
\end{figure}



The mesh coordinates $\mesh\subset\R^2$ are constructed such that the origin is the leading edge $\leadcoord{\vx}$ of the airfoil, that is, the leftmost point on $\surfmesh\subset\mesh$, or more formally
\begin{equation*}
    \leadcoord{\vx}\coloneq\argmin_{\vx\in\surfmesh}\compo=\vzero,
\end{equation*}
where $\compo$ denotes the first component of $\vx=\vect\compo\compl$. We can similarly define the trailing edge of the airfoil $\trailcoord\vx$ as the rightmost point on $\surfmesh$ by
\begin{equation*}
    \trailcoord{\vx}\coloneq\argmax_{\vx\in\surfmesh}\compo.
\end{equation*}
The mesh $\mesh$ can now be partitioned along the $\compo$-axis into three subsets of points as
\begin{equation*}
    \mesh=\freemesh\cup\foilmesh\cup\downmesh,
\end{equation*}
each of which are defined as
\begin{align*}
    \freemesh\coloneq\{\vx:\vx\in\mesh, \compo<\leadcoord{x}\} && \foilmesh\coloneq\{\vx:\vx\in\mesh, \compo\in[\leadcoord{x},\trailcoord{x}]\}
\end{align*}
\begin{equation*}
    \downmesh\coloneq\{\vx:\vx\in\mesh,x>\trailcoord\compo\}.
\end{equation*}
\todo{Add fig showing dynamics}
As can be seen in~\cref{fig:dynRegimes}, the dynamics occurring in each of these regions are distinct from one another. In the freestream region $\freemesh$, the behavior of the flow is largely independent of the airfoil. Upon entering $\foilmesh$, the airfoil surface deflects, compresses, and slows down air particles. These interactions are then carried downstream of the airfoil in the region described by $\downmesh$, where particles are still influenced by the airfoil geometry but no longer come into direct contact with its surface. It is therefore important for the model to be able to distinguish which partition a given mesh point belongs to for faithful modeling of the dynamics. Given the input representation $\innodefeatures\vx$ for the mesh point $\vx=\vect\compo\compl$ shown in~\cref{eq:basefeats}, it is straightforward for the model to determine whether $\vx\in\freemesh$ or $\vx\in\foilmesh\cup\downmesh$ via $\sign(x)$, as can be seen in~\cref{fig:leadingEdge}. However, distinguishing between $\foilmesh$ and $\downmesh$ is more challenging, as unlike $\leadcoord{\vx}$, $\trailcoord{\vx}$ can vary across airfoil geometries, for example according to the length $\trailcoord{x}-\leadcoord{x}$ of the airfoil. We therefore introduce a trailing edge coordinate system $\trailmesh\subset\R^2$ to better enable the model to distinguish between these regions. The trailing edge coordinate system, shown in~\cref{fig:trailingEdge}, places the origin at $\trailcoord\vx\in\mesh$, and is defined as
\begin{equation*}
\trailmesh\coloneq\{\vx-\trailcoord\vx:\vx\in\mesh\}.
\end{equation*}

Our \trailmodel model builds on the \svmodel model by adding the trailing edge coordinate system to the input node features as  
\begin{align*}
    \innodefeatures\vx=\cat{\basenodefeatures\vx, \trailnodefeatures\vx}\in\R^{11} && \trailnodefeatures\vx\coloneq\cat{\trailx\vx,\norm{\trailx\vx}}\in\R^3, 
\end{align*}
where $\trailx\vx\coloneq\vx-\trailcoord\vx\in\trailmesh$. Results for the \trailmodel model are presented in the following section.

\subsubsection{Hybrid Polar-Cartesian Coordinates}\label{sec:polarCart}

\begin{figure}
    \centering
    \includegraphics[width=0.5\linewidth]{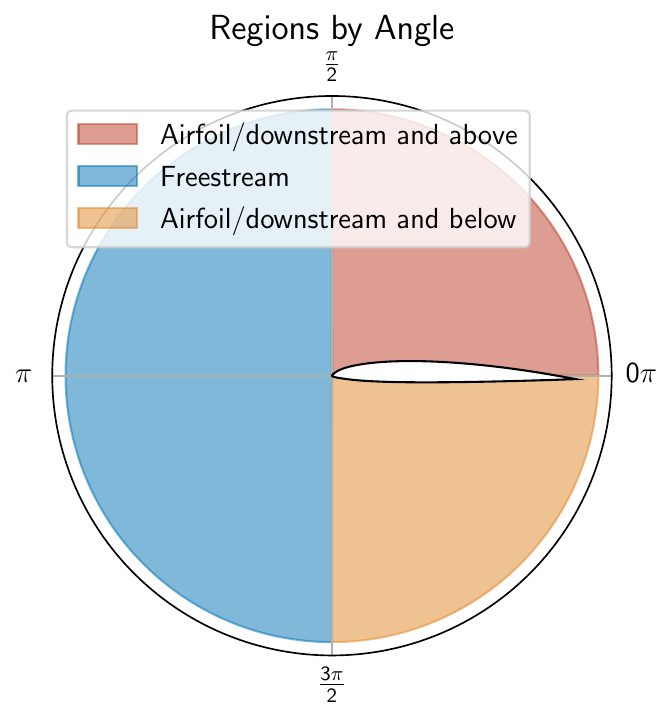}
    \caption{Regions by angle in the leading edge coordinate system. Polar angles can be used to derive both which of the regions along the $x$-axis defined in~\cref{fig:dynRegimes} a mesh point is in, as well as where the point is vertically with respect to the airfoil. The same information is contained in the Cartesian form of the leading edge coordinate system, although the model must learn interactions between the $\compo$ and $\compl$ coordinates in order to derive it.}
    \label{fig:angleRegion}
\end{figure}

\begin{figure}
    \centering
    \includegraphics[width=\linewidth]{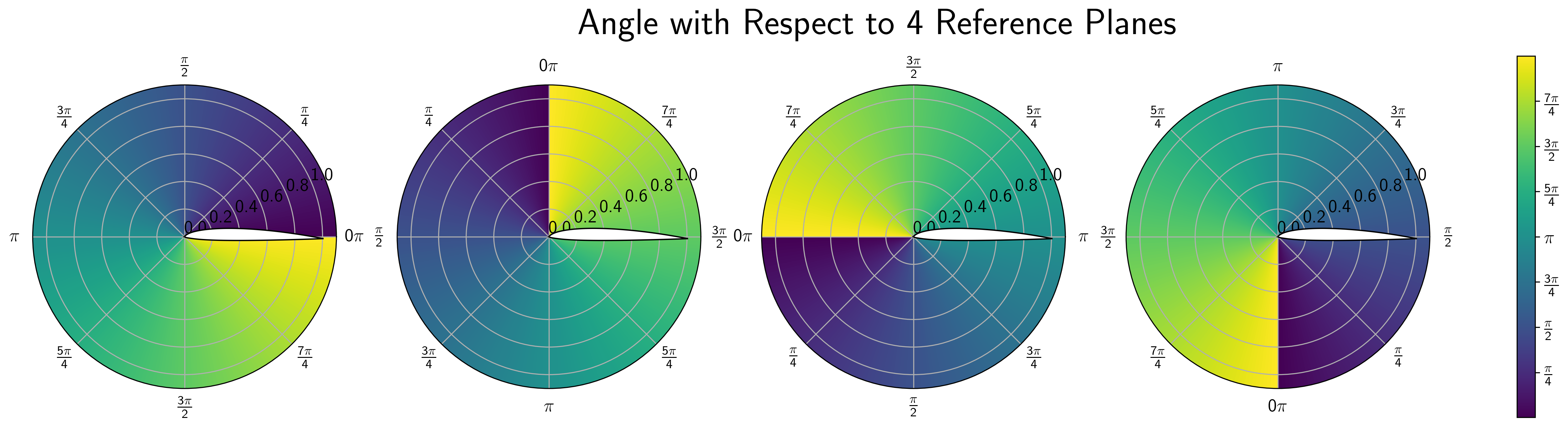}
    \caption{Angles in the leading edge coordinate system with respect to four different reference planes. Features in the \polarmodel model include angles with respect to all four reference planes.}
    \label{fig:fourAng}
\end{figure}

\begin{figure}
    \centering
    \includegraphics[width=\linewidth]{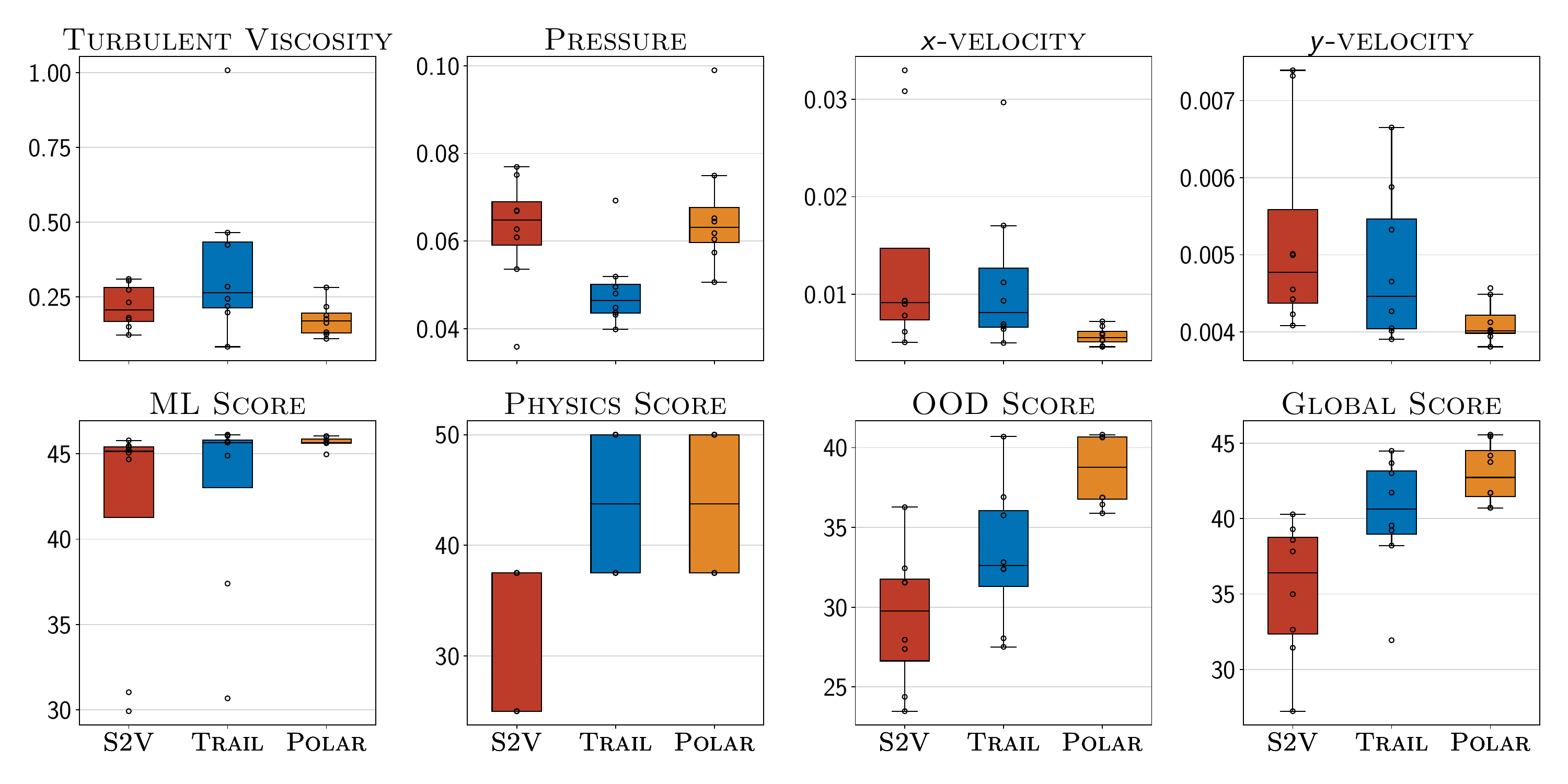}
    \caption{Comparison of the \trailmodel and \polarmodel models to the \svmodel model. }
    \label{fig:trailPolarRes}
\end{figure}

Although the \trailmodel model is able to distinguish between the regions defined in~\cref{fig:dynRegimes}, other distinctive regions are present which require the model to derive interactions between horizontal and vertical coordinates. For example, for $\vx=\vect{\compo}{\compl}\in\mesh$, $\sign(x)$ determines whether the mesh point is in $\freemesh$ or $\foilmesh\cup\downmesh$, while $\sign(y)$ indicates whether the mesh point is above or below the surface of the airfoil. If $\vx\in\freemesh$, then whether it is above or below the airfoil largely has no effect on the resultant dynamics, although if $\vx\in\foilmesh\cup\downmesh$, the dynamics above the airfoil are unique from those below, \textit{e.g.,} the generation of lift requires pressure on the lower surface of the airfoil to be greater than pressure on the upper surface~\citep{liu2021evolutionary}. 

To simplify learning, we extend model inputs to include the polar angle of $\vx=\vect\compo\compl$ with respect to polar axis $\vect10$ as $\ang\vx\coloneq\atan(y, x)$. As $\innodefeatures\vx$ also includes $\norm{\vx}$, the inputs can be seen to now contain both polar and Cartesian coordinates. Instead of learning interactions between coordinates, this enables derivation of relationships directly from $\ang\vx$. As shown in~\cref{fig:angleRegion}, this simplification can be seen through the previous example, where $\ang\vx\in(0,\frac\pi2]$ or $(\frac{3\pi}2,2\pi]$ indicates that $\vx$ is in $\foilmesh\cup\downmesh$ and above or below the airfoil, respectively, while $\ang\vx\in(\frac\pi2,\frac{3\pi}2]$ indicates that $\vx$ is in $\freemesh$. 

Instead of only using angle with respect to one reference plane, we compute angles with respect to 4 different polar axes $\vect10,\vect01,\vect{-1}0$ and $\vect0{-1}$ as shown in~\cref{fig:fourAng}. Angles are then computed as
\begin{equation*}
    \angf{\vx}\coloneq\cat{\ang\vx,\ang{\rtn{90}\vx},\ang{\rtn{180}\vx},\ang{\rtn{270}\vx}}\in\R^4, 
\end{equation*}
where $\rtn \nu$ denotes a rotation by $\nu$ degrees. Our \polarmodel model then extends the \trailmodel model by augmenting the input node features as 
\begin{align*}
    \innodefeatures\vx=\cat{\basenodefeatures\vx, \trailnodefeatures\vx, \angnodefeatures\vx}\in\R^{19} && \angnodefeatures\vx\coloneq\cat{\angf\vx,\angf{\trailx{\vx}}}\in\R^8. 
\end{align*}
Furthermore, the \polarmodel model additionally updates the input \sv edge features with the angle of the edge with respect to the polar axes as 
\begin{align*}
\inedgefeatures\vy\vx=\cat{\baseedgefeatures\vy\vx,\angedgefeatures\vy\vx}\in\R^7 && \angedgefeatures\vy\vx\coloneq\angf{\vy-\vx}\in\R^4.
\end{align*}

Results for the \trailmodel and \polarmodel models are presented in~\cref{fig:trailPolarRes}. Both the \trailmodel and \polarmodel models improve on the \score of the \svmodel model, with the largest gains coming from improvements in \physcore and \oodscore.

\subsection{Coordinate System Embedding}\label{sec:coordEmb}

\todo{\begin{itemize}
    \item Sinusoidal coordinate/ distance embeddings
    \item Spherical harmonics angle embeddings
\end{itemize}}
To enhance the expressiveness of input coordinates $x\in\R$, we expand them to $\cat{B_1(x),B_2(x),\ldots,B_{\nbasis}(x)}$, where $\{B_i\}_{i=1}^{\nbasis}$ is a $\nbasis$-dimensional basis. In the first linear layer $\mW$ of the embedding module, the model can then learn arbitrary functions of the coordinates $\sum_{i=1}^{\nbasis} w_iB_i(x)$. In~\cref{sec:sine}, we describe our embedding of Cartesian coordinates and distances using sinusoidal basis functions, while in~\cref{sec:sph}, we discuss spherical harmonic angle embeddings. In both cases, we take $\nbasis=8$.

\subsubsection{Sinusoidal Basis}\label{sec:sine}

\begin{figure}
    \centering
    \includegraphics[width=\linewidth]{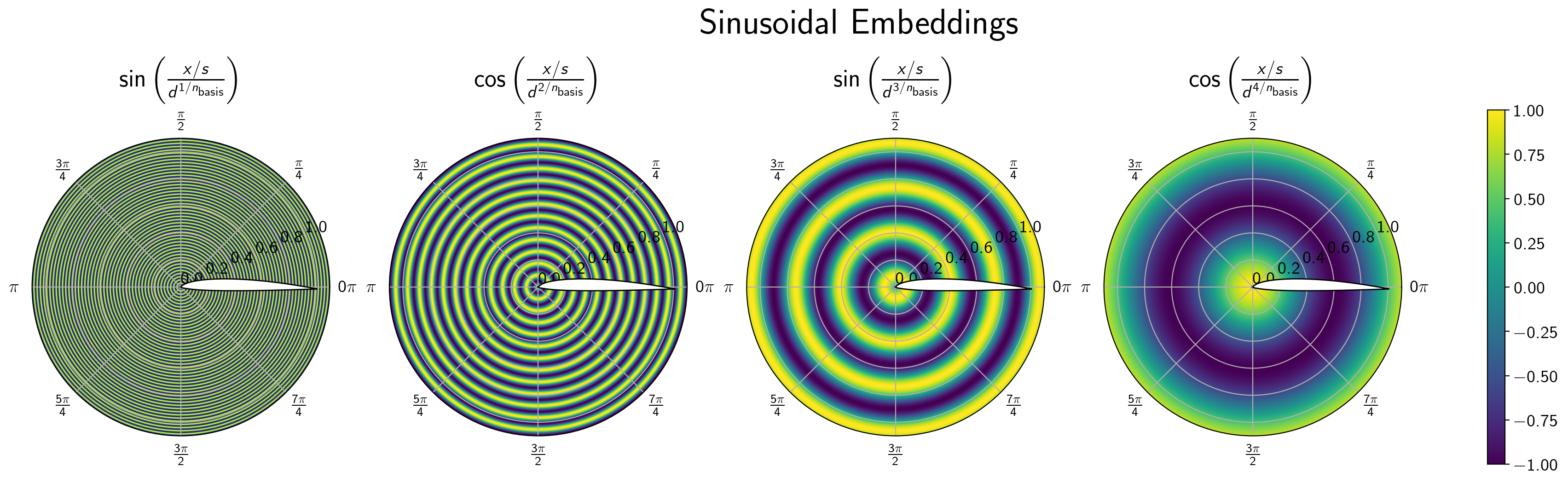}
    \caption{Sinusoidal embeddings of varying frequency for distance in the leading edge coordinate system.}
    \label{fig:sineEmb}
\end{figure}

For Cartesian coordinates and distances $x$, we employ the sinusoidal basis commonly used in Transformer models~\citep{attn} given by
\begin{equation}\label{eq:penc}
    \pe{x}\coloneq\left[\sin\left(\frac{x/s}{d^{i/\nbasis}}\right),\cos\left(\frac{x/s}{d^{i/\nbasis}}\right)\right]_{i=0}^{\nbasis-1}\in\R^{2\nbasis}.
\end{equation}
The division by $s$ in the denominator of~\cref{eq:penc} is to account for the spacing of the grid points, as \cite{attn} developed this scheme for $x\in\mathbb Z^{+}$ as a sequence position. While the spacing between consecutive sequence positions is $1$ for the setting considered by~\cite{attn}, for a computational mesh, the spacing between adjacent points can be substantially smaller. We furthermore set $d$
dependent on $s$ and the size of the domain $L$ as $d=\frac{4L}{s\pi}$. Sinusoidal embeddings for the distance in the leading edge coordinate system are shown in~\cref{fig:sineEmb}.

We form the \sinemodel model by augmenting the input node features of the \polarmodel model with sinusoidal embeddings of coordinates and distances as
\begin{align*}
        \innodefeatures\vx&=\cat{\basenodefeatures\vx, \trailnodefeatures\vx, \angnodefeatures\vx, \sinenodefeatures\vx}\in\R^{131} \\ \sinenodefeatures\vx&\coloneq\cat{\pe\vx,\pe{\trailx\vx},\pe{d(\vx)},\pe{\norm{\vx}}, \pe{\norm{\trailx\vx}}}\in\R^{112}, 
\end{align*}
where with a slight abuse of notation, we vectorize $\pefun$ as
\begin{equation}\label{eq:pemulti}
\pe\vx\coloneq\cat{\pe\compo,\pe\compl}\in\R^{4\nbasis}.
\end{equation}
Input \sv edge features are similarly augmented in the \sinemodel model with sinusoidal embeddings of edge displacements and distances as
\begin{align*}    
\inedgefeatures\vy\vx&=\cat{\baseedgefeatures\vy\vx,\angedgefeatures\vy\vx,\sineedgefeatures\vy\vx}\in\R^{55} \\ \sineedgefeatures\vy\vx&\coloneq\cat{\pe{\norm{\vy-\vx}},\pe{\vy-\vx}}\in\R^{48}.
\end{align*}
Results for the \sinemodel model are shown in the following section.


\subsubsection{Spherical Harmonics Basis}\label{sec:sph}

\begin{figure}
    \centering
    \includegraphics[width=\linewidth]{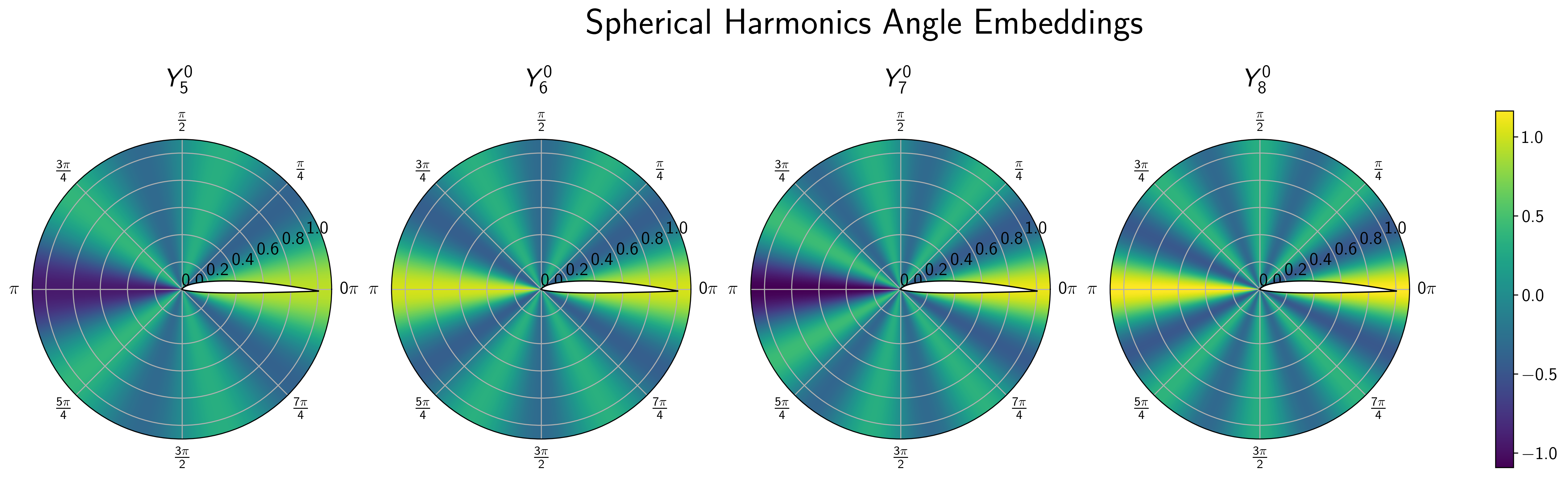}
    \caption{Spherical harmonics embeddings of angles in the leading edge coordinate system with $m=0$ and varying order $\ell$.}
    \label{fig:cos_sph}
\end{figure}

\begin{figure}
    \centering
    \includegraphics[width=\linewidth]{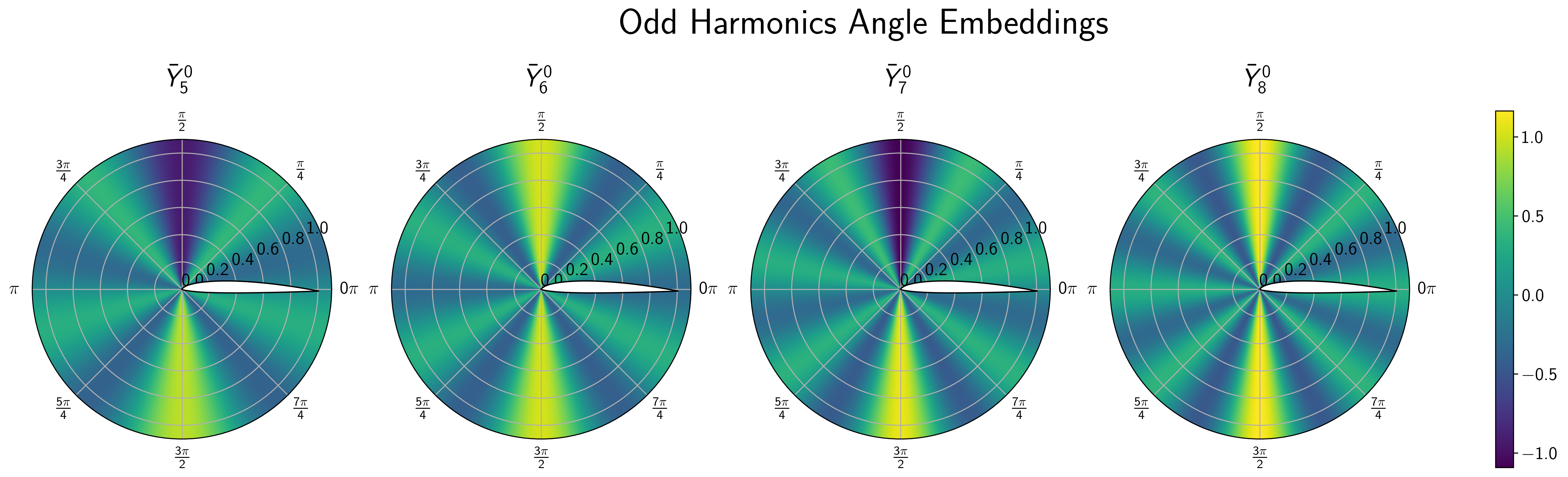}
    \caption{Odd harmonics embeddings of angles in the leading edge coordinate system with $m=0$ and varying order $\ell$.}
    \label{fig:sine_sph}
\end{figure}

\begin{figure}
    \centering
    \includegraphics[width=\linewidth]{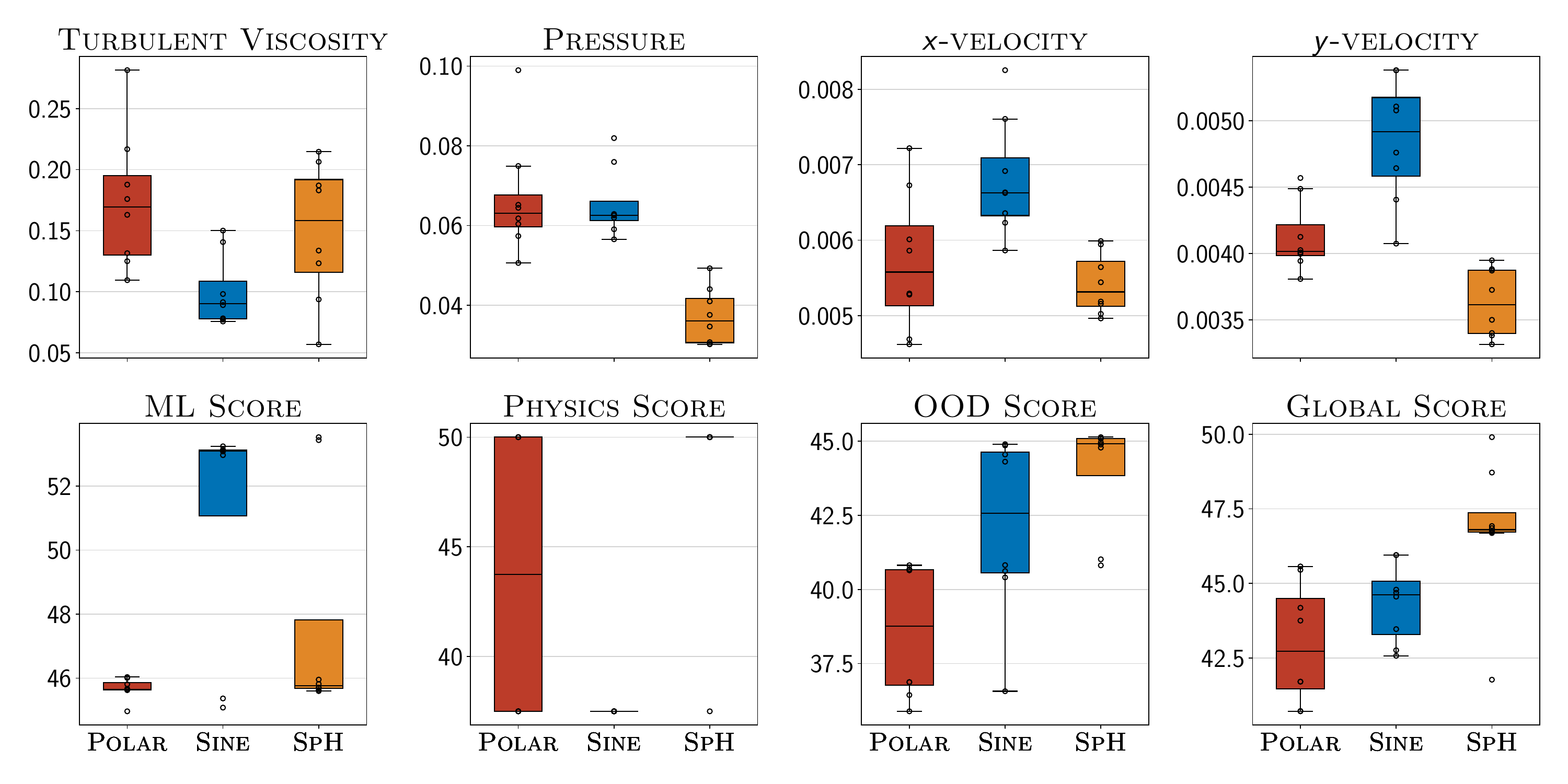}
    \caption{Comparison of the \sinemodel and \sphmodel models to the \polarmodel model.}
    \label{fig:sineSphhRes}
\end{figure}

For angles $\theta$, we instead use $m=0$ spherical harmonics embeddings~\citep{liu2022spherical,Gasteiger2020Directional}. The $m=0$ order $\ell$ spherical harmonic is given by~\citep{weisstein2004spherical}
\begin{equation}\label{eq:sph}
    \yl\ell\theta\coloneq\sqrt{\frac{(2\ell+1)!}{4\pi}}P_\ell(\cos(\theta))=\sum_{k=1}^\ell C_{\ell,k}\cos^k(\theta),
\end{equation}
where the coefficients $C_{\ell,k}\in\R$ are obtained from the order $\ell$ Legendre polynomial $P_\ell$. However, as can be seen in~\cref{eq:sph}, since $\yl\ell\theta$ is composed of powers of cosine, it is an even function, and as such, it cannot distinguish between $\theta$ and $-\theta$. We therefore define a set of odd basis functions $\oddyl\ell\theta$ as
\begin{equation}\label{eq:oddsph}
    \oddyl\ell\theta\coloneq\sqrt{\frac{(2\ell+1)!}{4\pi}}P_\ell(\sin(\theta)).
\end{equation}
As can be seen in~\cref{fig:sine_sph}, the odd harmonics can distinguish between $-\theta$ and $\theta$. We obtain embeddings of $\theta$ using both sets of bases in~\cref{eq:sph,eq:oddsph} as
\begin{equation*}
    \sph{\theta}\coloneq\left[\yl i\theta,\oddyl i\theta\right]_{i=1}^{\nbasis}\in\R^{2\nbasis}.
\end{equation*}
The \sphmodel model is formed by replacing angle features $\angnodefeatures\vx$ in the \sinemodel model with spherical harmonics embeddings of the angle features as 
\begin{align*}
        \innodefeatures\vx&=\cat{\basenodefeatures\vx, \trailnodefeatures\vx, \sphnodefeatures\vx, \sinenodefeatures\vx}\in\R^{251} \\ \sphnodefeatures\vx&\coloneq\cat{\sph{\angf\vx},\sph{\angf{\trailx\vx}}}\in\R^{128}, 
\end{align*}
where $\sphfun$ is vectorized as in~\cref{eq:pemulti} as
\begin{equation*}
\sph{\angf\vx}\coloneq\cat{\sph{\ang\vx},\sph{\ang{\rtn{90}\vx}},\sph{\ang{\rtn{180}\vx}},\sph{\ang{\rtn{270}\vx}}}\in\R^{8\nbasis}.     
\end{equation*}
\sv edge angles $\angedgefeatures\vy\vx$ are also replaced in the \sphmodel model as
\begin{align*}    \inedgefeatures\vy\vx&=\cat{\baseedgefeatures\vy\vx,\sphedgefeatures\vy\vx,\sineedgefeatures\vy\vx}\in\R^{115} \\ \sphedgefeatures\vy\vx&\coloneq\sph{\angf{\vy-\vx}}\in\R^{64}.
\end{align*}
Results for the \sinemodel and \sphmodel models are shown in~\cref{fig:sineSphhRes}. Both the \sinemodel and \sphmodel models improve on the \score and the \oodscore of the \polarmodel model. While the \sinemodel model achieves a better \mlscore due to improvements in the turbulent viscosity field error, the \sphmodel model achieves a new best error on the pressure field and improves the \physcore.  

\subsection{Turbulent Viscosity and Pressure Transformations}

In the previous sections, we have discussed techniques that we apply to all four models trained to predict velocities $\velx$ and $\vely$, pressure $\pres$, and turbulent viscosity $\visc$. However, as can be seen from the errors in~\cref{fig:sineSphhRes}, compared to velocity, the pressure and turbulent viscosity fields are particularly challenging to accurately model. We therefore introduce specific techniques for enhancing model generalization on each of these fields. In~\cref{sec:canon}, we derive a change of basis to canonicalize input features with respect to inlet velocity. We additionally re-parameterize our model to predict the log-transformed pressure, as discussed in~\cref{sec:logPres}.   

\subsubsection{Inlet Velocity Canonicalization}\label{sec:canon}

\begin{figure}
    \centering
    \includegraphics[width=\linewidth]{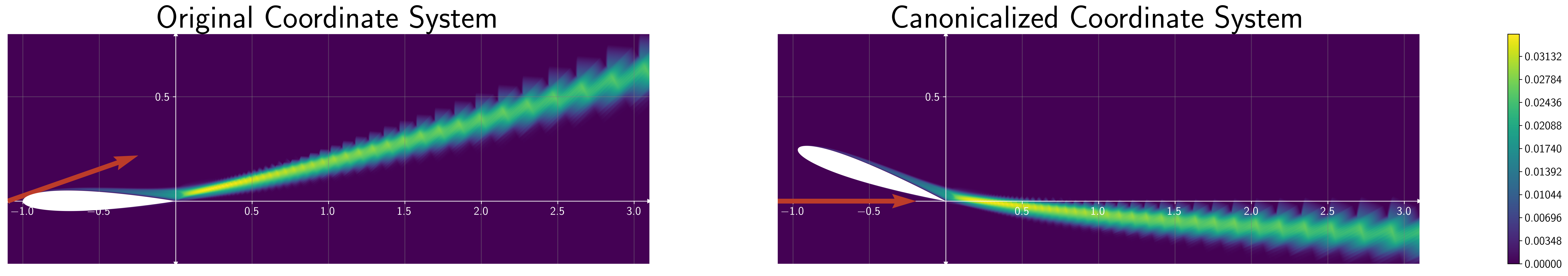}
    \caption{Turbulent viscosity and inlet velocity in the leading edge coordinate system (\textit{left}) and in the canonicalized leading edge coordinate system (\textit{right}). The direction of the inlet velocity, shown in red, is strongly associated with the direction of the non-zero parts of the turbulent viscosity field. After canonicalizing the coordinate system such that the inlet velocity is parallel with the $x$-axis, the non-zero parts of the turbulent viscosity are concentrated in the region around the $x$-axis for all examples.}
    \label{fig:canonTV}
\end{figure}

\begin{figure}
    \centering
    \includegraphics[width=\linewidth]{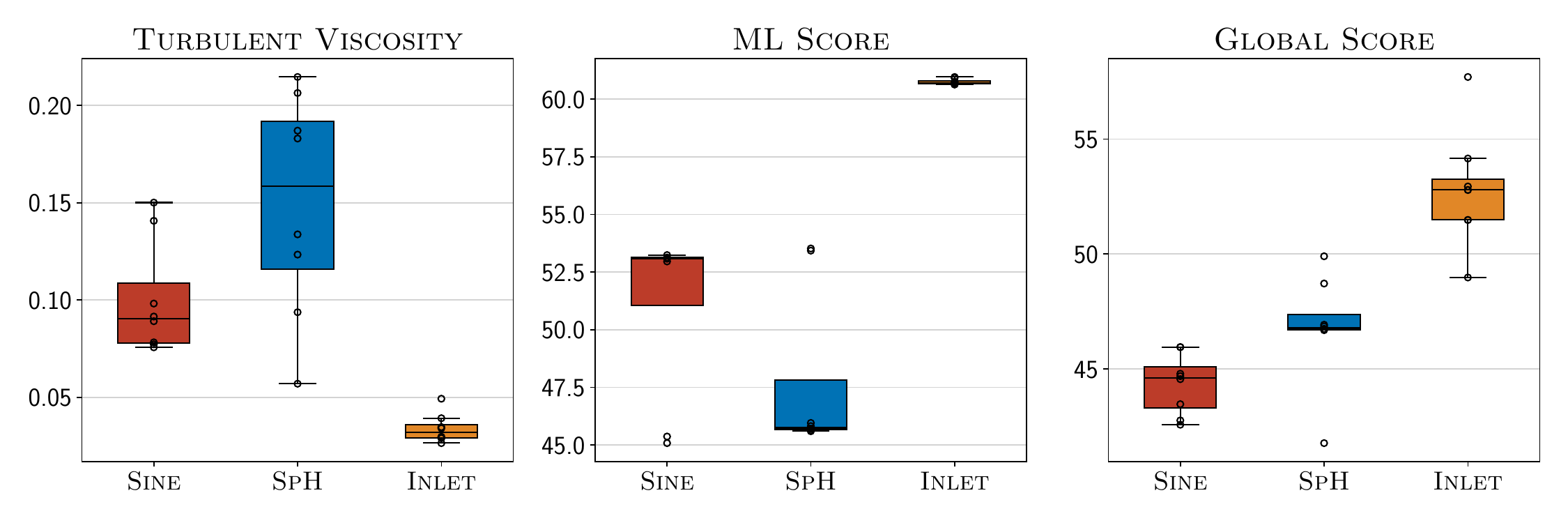}
    \caption{Comparison of the \canonmodel model to the \sinemodel and \sphmodel models, which previously had the best error on turbulent viscosity and the best \score, respectively.}
    \label{fig:canonTVres}
\end{figure}

\todo{Make sure to mention that we only do this for pressure and viscosity}

As can be seen in~\cref{fig:canonTV}, the non-zero regions of the turbulent viscosity $\visc$ are strongly associated with the direction of the inlet velocity $\inletvel$. To improve generalization on this field, we apply a change-of-basis to obtain a new coordinate system under which $\inletvel$ is rotated to be parallel with the $\compo$-axis. This effectively \textit{canonicalizes} the coordinate system with respect to the inlet velocity, a technique which has been applied in machine learning to improve generalization~\citep{puny2022frame,lin2024equivariance}. In the case of $\visc$, as can be seen in~\cref{fig:canonTV}, the association between $\inletvel$ and $\visc$ implies that many of the non-zero parts of $\visc$ will occur in the region along the $\compo$-axis in the canonicalized coordinate system, resulting in a substantially less difficult modeling task.

This canonicalization can be achieved with the rotation matrix $\inrtn\in O(2)$ given by
\begin{equation*}
    \inrtn\coloneq\frac1{\norm{\inletvel}} 
    \begin{bmatrix}
        v_1 & v_2
        \\
        -v_2 & v_1
    \end{bmatrix},
\end{equation*}
where $\inletvel=\vect{v_1}{v_2}$. $\inrtn$ is obtained by deriving the matrix that rotates $\inletvel$ to be parallel with the $\compo$-axis, that is
\begin{equation*}
    \inrtn\inletvel=\vect{\norm{\inletvel}}0.
\end{equation*}
For the models trained to predict turbulent viscosity and pressure, we extend the input node features for the \canonmodel model from those for the \sphmodel model to additionally include coordinates, angles, and surface normals in the canonicalized coordinate system as
\begin{align*}
    \innodefeatures\vx&=\cat{\basenodefeatures\vx, \trailnodefeatures\vx, \sphnodefeatures\vx, \sinenodefeatures\vx, \canonodefeatures\vx}\in\R^{449} \\ \canonodefeatures\vx&\coloneq\cat{\inrtn\vx, \inrtn\trailx{\vx}, \pe{\inrtn\vx}, \pe{\inrtn\trailx{\vx}}, \inrtn\vn(\vx),\sph{\angf{\inrtn\vx}}, \sph{\angf{\inrtn{\trailx\vx}}}}, 
\end{align*}
with $\canonodefeatures\vx\in\R^{198}$. The \canonmodel model adds the canonicalized \sv edge displacements and angles in the input edge features as 
\begin{align*}    \inedgefeatures\vy\vx&=\cat{\baseedgefeatures\vy\vx,\sphedgefeatures\vy\vx,\sineedgefeatures\vy\vx,\canonedgefeatures\vy\vx}\in\R^{213} \\ \canonedgefeatures\vy\vx&\coloneq\cat{\inrtn(\vy-\vx),\pe{\inrtn(\vy-\vx)},\sph{\angf{\inrtn(\vy-\vx)}}}\in\R^{98}.
\end{align*}
Results for the \canonmodel model are shown in~\cref{fig:canonTVres}. We compare to the \sinemodel model, which previously had the best error on the turbulent viscosity field, and the \sphmodel model, which previously had the best \score. The \canonmodel model improves both metrics.  

\subsubsection{Log-Transformed Pressure Prediction}\label{sec:logPres}

\begin{figure}
    \centering
    \includegraphics[width=\linewidth]{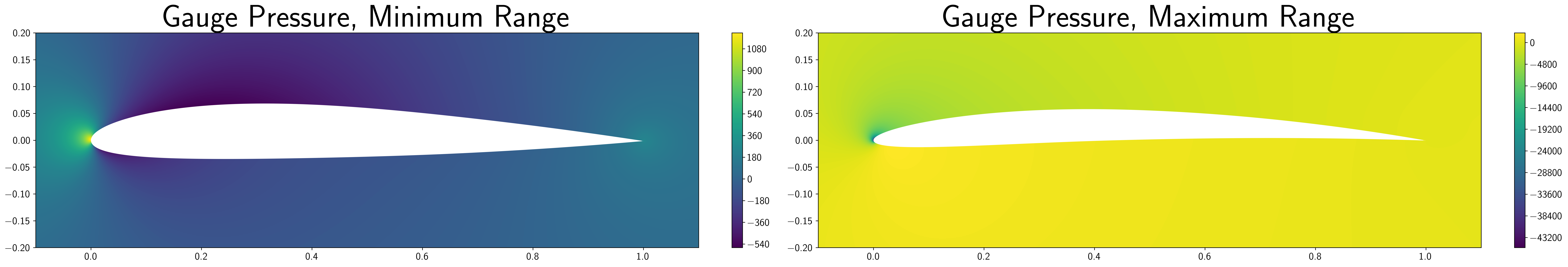}
    \caption{Pressure field with the least range (\textit{left}) and with the greatest range (\textit{right}). }
    \label{fig:press}
\end{figure}


\begin{figure}
\centering
\begin{subfigure}{.5\textwidth}
  \centering
  \includegraphics[width=\linewidth]{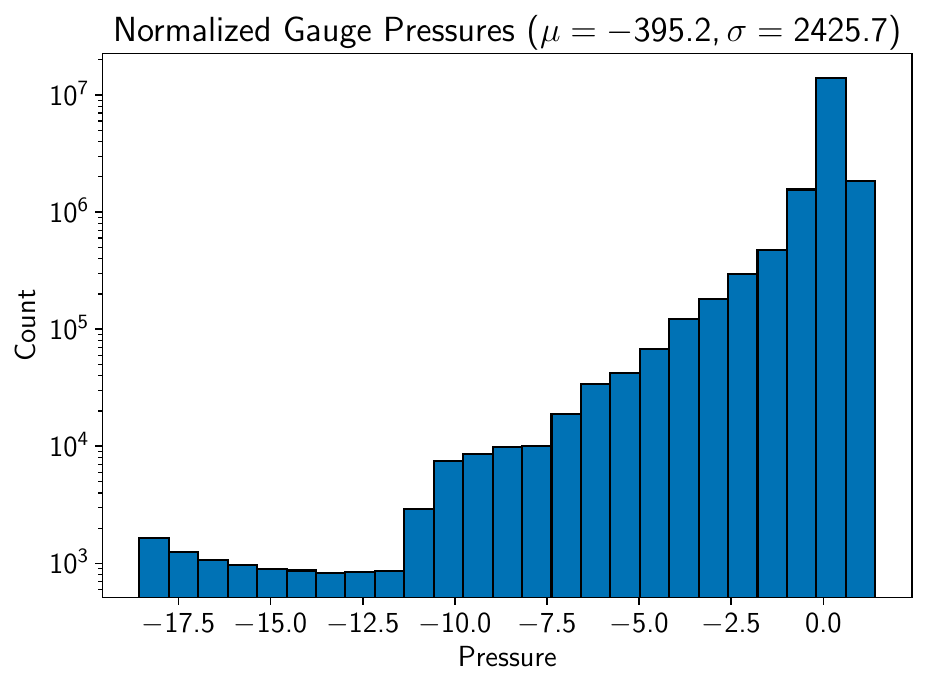}
  \caption{Pressure.}
  \label{fig:pressureDist}
\end{subfigure}%
\begin{subfigure}{.5\textwidth}
  \centering
  \includegraphics[width=\linewidth]{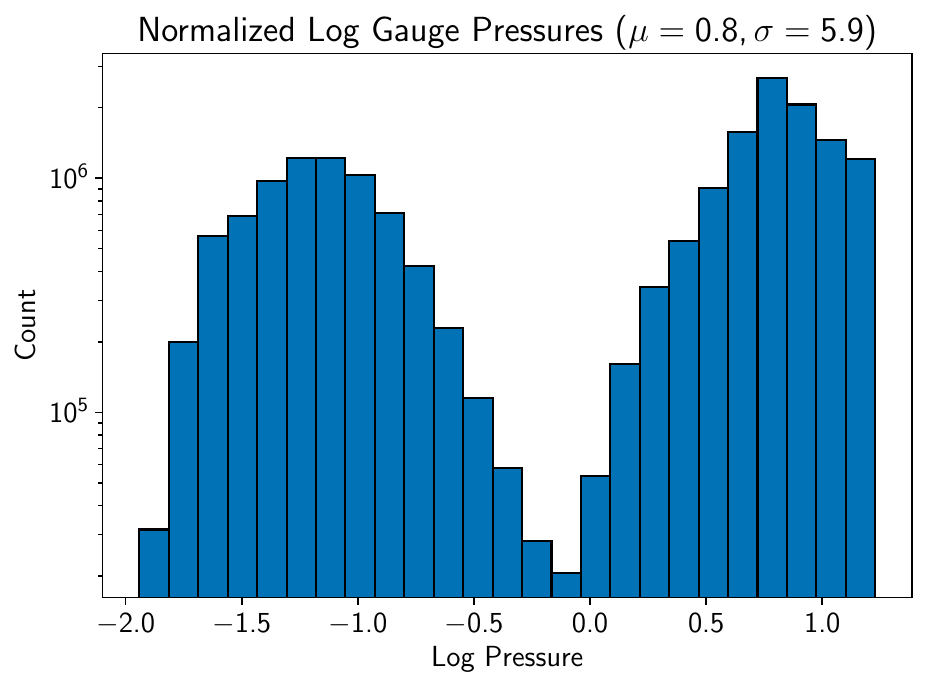}
  \caption{Log-transformed pressure.}
  \label{fig:logPressureDist}
\end{subfigure}
\caption{Distribution of normalized pressures.}
\label{fig:pressureDists}
\end{figure}

\begin{figure}
    \centering
    \includegraphics[width=\linewidth]{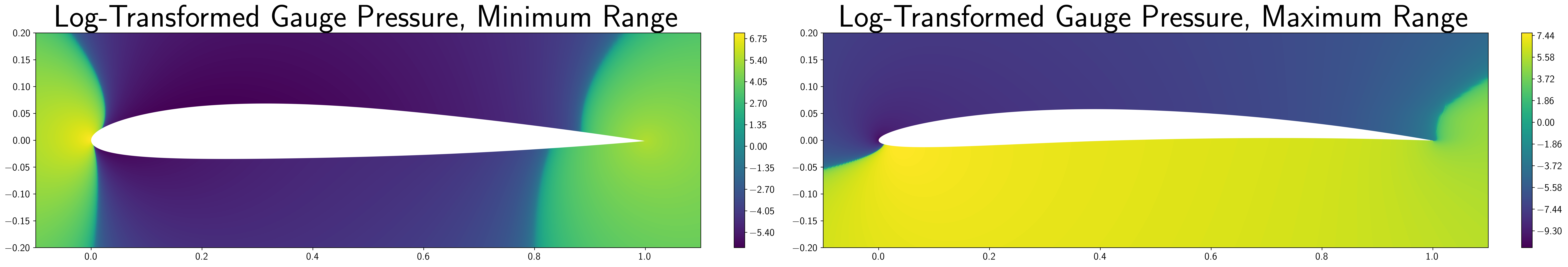}
    \caption{Log-transformed pressure field with the least range (\textit{left}) and with the greatest range (\textit{right}).}
    \label{fig:logPres}
\end{figure}

\begin{figure}
    \centering
    \includegraphics[width=\linewidth]{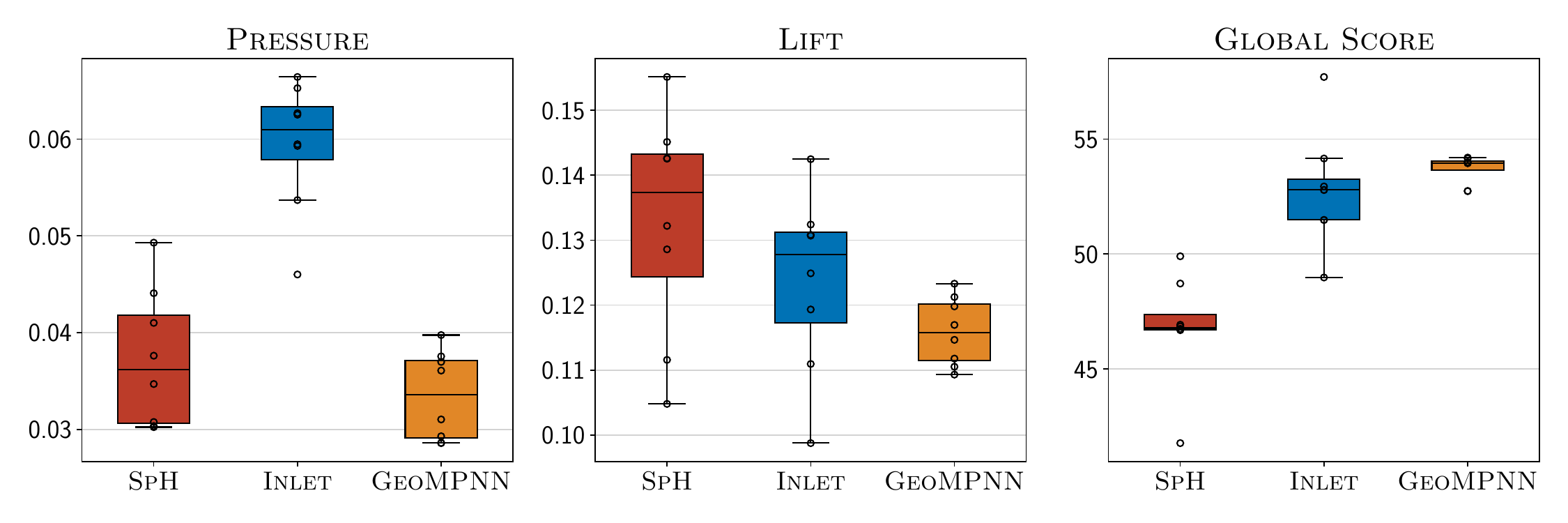}
    \caption{Comparison of the \geompnnmodel model to the \sphmodel and \canonmodel models, which previously had the best error on pressure and the best \score, respectively.}
    \label{fig:logPresRes}
\end{figure}




Lift forces acting on the airfoil are primarily generated through the difference between pressure on the lower and upper surfaces of the airfoil~\citep{liu2021evolutionary}. Therefore, accurate modeling of the pressure $\pres$ is necessary for accurate prediction of the lift coefficient $\lift$. However, this large differential leads to a more challenging prediction target. As can be seen in~\cref{fig:press,fig:pressureDist}, there is a large amount of variance between the low pressure region above the airfoil and the high pressure region below the airfoil. Note that $\pres(\vx)$ is the \textit{gauge pressure}, which represents the deviation of the absolute pressure 
$p_{\operatorname{abs}}$ from the freestream pressure $p_\infty$ as
\begin{equation}
    \pres(\vx)\coloneq p_{\operatorname{abs}}(\vx) - p_\infty.
\end{equation}
While log transformations are often used to reduce the variance in positive-valued data, the gauge pressure can also take negative values, and thus, we introduce a log-transformation of the pressure as
\begin{equation*}
    \logpres(\vx)\coloneq\sign(\pres(\vx))\log(\lvert\pres(\vx)\rvert+1).
\end{equation*}
As can be seen in~\cref{fig:logPressureDist,fig:logPres}, this substantially reduces variance. Our final \geompnnmodel model modifies the training procedures for the \canonmodel model to predict the log-transformed pressure $\logpres(\vx)$. The untransformed pressure $\pres$ can then be obtained from $\logpres$ by applying the inverse transformation given by
\begin{equation*}
    \pres(\vx)=\sign(\logpres(\vx))(\exp(\lvert\logpres(\vx)\rvert) - 1).
\end{equation*}

Results for the \geompnnmodel model are shown in~\cref{fig:logPresRes}. We compare to the \sphmodel model, which previously had the best error on the pressure field, and the \canonmodel model, which previously had the best \score. The \geompnnmodel model improves both metrics, and also achieves a lower relative error for the lift coefficient.

\section{Conclusion}

We propose \geompnnmodel to accelerate the steady-state solution of the RANS equations over airfoils of varying geometries subjected to different simulation conditions. Integral to the architecture of \geompnnmodel is the incorporation of an expressive geometric representation of the airfoil in model predictions using \propmp. We furthermore demonstrate that the design of \geompnnmodel enables efficient training on subsampled meshes with no degradation in prediction quality on the full mesh. We design physically-inspired coordinate system representations and a canonicalization scheme to enhance the generalization capabilities of \geompnnmodel. Extensive experiments validate our design choices and show that \geompnnmodel can accurately and efficiently model dynamics for a range of airfoil geometries and simulation conditions.


\section*{Acknowledgments}
This work was supported in part by National Science Foundation under grant IIS-2243850.

\bibliography{bib}

\end{document}